\pdfoutput=1

\documentclass[11pt]{article}

\usepackage[]{acl}

\usepackage{times}
\usepackage{latexsym}
\usepackage{amsmath}
\usepackage{dsfont}
\usepackage{graphicx}
\usepackage{subcaption}
\usepackage{multirow}
\newcommand\scalemath[2]{\scalebox{#1}{\mbox{\ensuremath{\displaystyle #2}}}}

\usepackage[T1]{fontenc}

\usepackage[utf8]{inputenc}

\usepackage{microtype}

\title{The Craft of Selective Prediction: Towards Reliable \\ Case Outcome Classification - An Empirical Study on \\ European Court of Human Rights Cases}

\author{ Santosh T.Y.S.S, \bf{Irtiza Chowdhury, Shanshan Xu, Matthias Grabmair} \\ School of Computation, Information, and Technology; \\
Technical University of Munich, Germany \\ \ }

\begin{document}
\maketitle
\begin{abstract}
In high-stakes decision-making tasks within legal NLP, such as Case Outcome Classification (COC), quantifying a model's predictive confidence is crucial. Confidence estimation enables humans to make more informed decisions, particularly when the model's certainty is low, or where the consequences of a mistake are significant. However, most existing COC works prioritize high task performance over model reliability. This paper conducts an empirical investigation into how various design choices—including pre-training corpus, confidence estimator and fine-tuning loss—affect the reliability of COC models within the framework of selective prediction. Our experiments on the multi-label COC task, focusing on European Court of Human Rights (ECtHR) cases, highlight the importance of a diverse yet domain-specific pre-training corpus for better calibration. Additionally, we demonstrate that larger models tend to exhibit overconfidence, Monte Carlo dropout methods produce reliable confidence estimates, and confident error regularization effectively mitigates overconfidence. To our knowledge, this is the first systematic exploration of selective prediction in legal NLP. Our findings underscore the need for further research on enhancing confidence measurement and improving the trustworthiness of models in the legal domain.

\end{abstract}

\section{Introduction}
The task of Case Outcome Classification (COC) involves categorizing the outcome of a legal case based on the text of case facts. It has garnered substantial attention not only within the legal but also in the NLP community \cite{aletras2016predicting, chalkidis2019neural, hwang2022multi}
. It is important to acknowledge that these tasks commonly labeled as `Legal Judgment Prediction' are, in reality, instances of retrospective classification rather than prediction as the fact statements obtained from judgment documents are often not finalized until the decision outcome is known, as emphasized by \citealt{medvedeva2021automatic}, introducing the potential confounding artifacts in data \cite{santosh2022deconfounding}. 
The main utility of this task and data lies in understanding the capabilities of NLP models to analyze fact statements for extracting and learning text patterns corresponding to specific convention articles, as drafted by the court.  Identifying potentially violated human rights provisions from a textual fact description is a task that human experts can do well and that requires substantial domain knowledge along with textual understanding phenomenon \cite{chalkidis2022lexglue}. 
Correlation between text patterns and violations has been a subject of interest in empirical research, particularly in political science \cite{segal1984predicting, kort1957predicting, nagel1963applying}.

Though COC has witnessed improvements in performance with use of better pre-trained models \cite{douka2021juribert,chalkidis2020legal,xiao2021lawformer} or innovative modelling strategies with better architectures and loss functions \cite{tyss2023leveraging,yue2021neurjudge,zhao2022cpee,zhong2018legal} or incorporation of legal knowledge \cite{tyss2023zero,gan2021judgment}, the accuracy of these models is not guaranteed for all instances. Therefore, it is crucial to assess the reliability of model predictions, particularly in high-stakes decision-making scenarios—an aspect that has not received adequate attention in the community. 

In this paper, we explore the reliability of COC systems using selective prediction setting \cite{el2010foundations}. In this setting, the objective is to reduce the error rate by abstaining from predictions when the model is uncertain, while maintaining high coverage. In essence, we consider a model reliable if it possesses the self-awareness capability to acknowledge when it doesn’t know enabling it to defer to humans for manual inspection, thereby ensuring its trustworthiness \cite{geifman2017selective}. Under the selective prediction setting, we construct a selective classifier by combining a standard classifier with a confidence estimator. The confidence estimator gauges the model's confidence for a given input instance and based on this confidence, the selective classifier decides whether to abstain from predicting on uncertain cases. An ideal confidence estimator should provide higher confidence for correctly classified examples compared to incorrect ones.

In this study, we conduct an empirical investigation on the COC task, focusing on European Court of Human Rights (ECtHR) cases which adjudicates complaints by individuals against states regarding alleged violations of their rights as enshrined in the European Convention of Human Rights. Our goal is to assess four design choices to obtain more reliable COC models: (i) How does the choice of pre-trained models, such as general BERT \cite{kenton2019bert} or legal domain-specific models like LegalBERT \cite{chalkidis2019neural} or LexLM \cite{chalkidis2023lexfiles}, impact reliability? Does the size of the pre-trained model, such as Base or Large, play a role? (ii) Is there a universally effective confidence estimator, such as Softmax Response \cite{hendrycks2016baseline} or Monte Carlo dropout based methods\cite{gal2016dropout}? 
(iii) How do additional training loss constraints in the form of regularizers \cite{xin2021art,shamsiuncertainty} or learning directly with abstention as an option \cite{liu2019deep} affect it? We evaluate these design choices on three COC task variants with varying difficulty, ranging from predicting violations alleged by the claimant to violations decided by the court. Additionally, we examine the impact of these choices on different buckets of articles based on frequency of cases with corresponding article violations.

Based on our empirical exploration, we observe: (i) Domain-specific pre-training enhances model calibration, but exclusive focus on downstream task-specific pre-training corpus is detrimental. A domain-related yet diverse corpus is crucial for effective pre-training. Larger models, while more accurate, tend to be overconfident. (ii) Computationally expensive Monte Carlo Dropout methods provide superior confidence estimates. (iii) Adding confident error regularization \cite{xin2021art} improves model calibration. To encourage future work towards better uncertainty quantification in COC task, we release our code, including pipelines to evaluate design choices based on selective prediction and classification performance.

\section{Related Work}
\noindent \textbf{Selective Prediction} Selective prediction, in which a model can either predict or abstain on each test example, is a long-standing research area in machine learning \cite{chow1957optimum,hellman1970nearest,fumera2002support,cortes2016learning,el2010foundations,geifman2017selective}. Selective prediction has recently received considerable attention from the NLP community on various tasks such as Quesrion answering \cite{kamath2020selective,garg2021will}, classification and NLI \cite{gu2023evaluation,varshney2022towards,varshney2022investigating}, knowledge probing \cite{yoshikawa2023selective} and generation \cite{ren2022out,chen2023adaptation,cole2023selectively} and is mostly related to uncertainty/confidence estimation \cite{vazhentsev2022uncertainty,vazhentsev2023hybrid}. Another related area to selective prediction, albeit remotely, is calibration \cite{jiang2018trust,desai2020calibration,wang2020inference,guo2017calibration} which deals with the development of interpretable confidence measures focusing on adjusting the overall confidence level of a model, while selective prediction is based on relative confidence among the examples. In this work, we employ selective prediction framework to evaluate the reliability of models in the context of multi-label COC task, in contrast to prior works that predominantly focused on single-label classification tasks.

\noindent \textbf{COC} 
COC has been explored using corpora from different jurisdictions, such as the ECtHR \cite{chalkidis2022lexglue, aletras2016predicting, medvedeva2021automatic,santosh2023Zero,santosh2023leveraging} 
Chinese Criminal Courts \cite{ yue2021neurjudge}
, US Supreme Court \cite{katz2017general,kaufman2019improving}, Indian Supreme Court \cite{malik2021ildc,shaikh2020predicting}
French court of Cassation \cite{csulea2017predicting}, 
Federal Supreme Court of Switzerland \cite{niklaus2021swiss}, 
UK courts \cite{strickson2020legal}, German courts \cite{waltl2017predicting}, Brazilian courts \cite{lage2022predicting}, the Philippine Supreme court \cite{virtucio2018predicting}, and the Thailand Supreme Court \cite{kowsrihawat2018predicting}. 
While early works relied on rule-based approaches \cite{segal1984predicting,nagel1963applying}, 
later works used classification techniques using bag-of-words features \cite{aletras2016predicting, csulea2017exploring}. 
Most recent work in COC use deep learning \cite{zhong2018legal, zhong2020iteratively, yang2019legal} followed by adoption of pre-trained transformer models \cite{chalkidis2019neural,niklaus2021swiss}, including legal-domain specific pre-trained variants \cite{zheng2021does,chalkidis2023lexfiles}. 
Furthermore, different strategies were proposed by leveraging dependency between auxiliary tasks \cite{tyss2023leveraging,yue2021neurjudge,valvoda2023role} 
or with additional loss such as contrastive learning \cite{tyss2023leveraging,zhang2023contrastive} 
or by injecting legal knowledge \cite{liu2023ml,tyss2023zero}. 

While the majority of existing research in COC focuses on improving predictive performance, there is an increasing emphasis on the reliability of models within legal NLP. This includes perspectives on explainability \cite{chalkidis2021paragraph, santosh2022deconfounding, xu2023dissonance} and fairness \cite{wang2021equality, chalkidis2022fairlex, li2022fairness,baumgartner2024towards}. Recently, mainstream NLP has seen researchers, such as \citealt{baan2024interpreting}, propose that the overall reliability of a model is determined by two facets: 1) fairness, which is related to the alignment of model confidence with human expectations, and 2) trustworthiness, which involves accurate confidence measurement by the model. \citealt{xu2024through} study models' calibration in alignment with human behavior using split-vote cases from ECtHR, pioneering on research on model reliability from a fairness perspective in legal NLP. In contrast, our work systematically explores selective prediction in COC task. To the best of our knowledge, this is the first COC research on model reliability from a trustworthiness perspective in Legal NLP. 


\section{ECtHR Tasks \& Datasets}
We chose to work with the ECtHR corpus because of its publicly available dataset with detailed article-specific allegations and violation information, leading to multi-label classification setting, in contrast to the simplified binary classification setting in corpora from other jurisdictions \cite{niklaus2021swiss, alali2021justice}. 
We experiment with the following three COC task variants. Following \citealt{valvoda2023role}, we use the 14 articles which form the core rights of the convention. 

\noindent \textbf{Task B: Allegation Identification} \cite{chalkidis2021paragraph} We utilize data from LexGLUE ECtHR B \cite{chalkidis2022lexglue}, where the fact description serves as input to identify the set of convention articles that the claimant alleges to have been violated.

\noindent \textbf{Task A: Violation Identification} \cite{chalkidis2019neural} We leverage data from LexGLUE ECtHR A to predict which of the convention’s articles has been deemed violated by the court using the facts description as input. Task A is more challenging than B as it involves the identification of suitable articles along with prediction of their violations.

\noindent \textbf{Task A|B: Violation Identification given Allegation information} \cite{santosh2022deconfounding2} This involves the identification of violations from the case facts along with allegedly violated articles as the input. This task mirrors the realistic legal process, as the court is aware of the allegations made by the applicants when determining the violations. This task is easier compared to Task A, as the first sub-step of Task A (i.e., identifying suitable articles) is provided directly as input in this variant.

\noindent \textbf{Dataset splits \& Metrics for Prediction Performance}  LexGLUE consists of of 11k case fact descriptions chronologically split into training (2001–2016, 9k cases), validation (2016–2017, 1k cases), and test sets (2017-2019, 1k cases). Following \citealt{chalkidis2022lexglue}, we report 
macro-F1 (m-F1) scores for all tasks across the 14 articles. 

\section{Selective Prediction}
A standard classifier learns a function $f:X \rightarrow Y$, takes input $X$ and maps it to set of labels $Y$. We pair a standard classifier with a selection function $g:X \rightarrow \{0, 1\}$ to obtain a selective classifier $h =(f, g)$; $h:X\rightarrow Y \cup \{\perp\}$, $\perp$ is a special label indicating the abstention of prediction. Given an input $x$, the selective classifier outputs as follows:
\[
h(x) = (f, g)(x) = 
\begin{cases}
  f(x) & \text{if $g(x)=1$} \\
  \perp & \text{if $g(x)=0$}
\end{cases}
\]
The selective classifier yields an output from $f$ when the selection function predicts that prediction should be given, or abstains if the selection function predicts that it should not predict. Convenient way to formulate the selection function $g$ is relying on a confidence function $\tilde{g}$ and a threshold $\gamma \in \mathds{R}$ as:
\begin{equation}
    g(x) = \mathds{1}[\tilde{g}(x) > \gamma]
\label{threshold}
\end{equation}
where confidence function $\tilde{g}: X \rightarrow \mathds{R}$ assigns a real-valued confidence to an instance $x \in X$. Ideally, a good confidence estimator $\tilde{g}(x)$ for abstention should yield high values when $f(x)$ is correct and low values when it is incorrect.

\noindent  \textbf{Metrics for Selective Prediction} Coverage ($C$) is the portion of instances that the model choose to predict, while risk ($R$) is the error on that subset of predictions. For a selective classifier $h=(f,g)$ on dataset $D$ with inputs $x_i$ and ground truth labels $y_i$, they are given as follows:
\begin{equation}
    C(h)= \frac{1}{|D|} \sum\limits_{(x_i,y_i) \in D} g(x_i)
\end{equation}
\begin{equation}
    R(h) = \frac{\frac{1}{|D|} \sum\limits_{(x_i,y_i) \in D} l(f(x_i),y_i)·g(x_i) }{C(h)}
\end{equation}
where loss function $l$ measures the error between the predicted label $f(x_i)$ and the ground truth $y_i$. Ideally, a reliable model should showcase high coverage at low levels of risk, implying accurate predictions for many instances and abstention on others. As the threshold $\gamma$ in Eq. \ref{threshold} decreases, coverage increases, but risk also rises. Hence there exists a risk-coverage trade-off that models strive to optimize. Thus, we construct a curve plotting coverage versus the corresponding risk \cite{el2010foundations} and calculate the Area Under Risk-Coverage Curve (AURCC). A lower AURCC indicates a better selective classifier.

\noindent We calculate Reversed Pair Proportion (RPP) following \citealt{xin2021art}, a normalized version of the Kendall-Tau distance \cite{kendall1948rank} to gauge how closely the confidence estimator  $\tilde{g}$ aligns with the ideal. Ideally, the confidence estimator should rank all incorrect predictions below all correct predictions. RPP quantifies the proportion of instance pairs with a reversed confidence–error relationship.
\begin{equation}
    RPP = \frac{\sum\limits_{1 \leq i,j \leq |D|} \mathds{1}[\tilde{g}(x_i) < \tilde{g}(x_j), l_i < l_j ] }{|D|^2}
\end{equation}
A lower RPP value indicates better estimator. However, \citealt{gu2023evaluation} highlights that RPP and AUC values are influenced by both the prediction and confidence estimation functions. 
They propose refinement metric, normalizing with the worst-case Kendall-tau distance, 
offering a calibrated interpretable metric where 0, 0.5 and 1 signifies the best, random and the worst case respectively. 
\begin{equation}
    Rf =  \frac{\sum\limits_{1 \leq i,j \leq |D|} \mathds{1}[\tilde{g}(x_i) < \tilde{g}(x_j), l_i < l_j ] }{c(|D|-c)}
\end{equation}
where $c$ denote number of correct predictions made by the prediction function. 

\subsection{Confidence Estimators}
\label{confidence}
\noindent \textbf{Softmax Response (SR)} \cite{hendrycks2016baseline} derives the confidence estimate based on the maximum probability assigned to one of the labels. 
\begin{equation}
    \tilde{g}_{SR}(x) = \max\limits_{y \in Y} p(y)
\end{equation}

\noindent \textbf{Monte Carlo (MC) Dropout} \cite{gal2016dropout} Different dropout value is used to derive the confidence estimate of a neural network by computing $p(y)$ for a total of N times 
and we consolidate these N probability values into a confidence estimate using the below three variants.

\noindent  \emph{Sampled maximum probability} (SMP) uses the sample mean as the final confidence. $p_y^n$ is the probability of the label $y$ obtained using the $n^{th}$ mask. 
 
\begin{equation}
    \tilde{g}_{SMP}(x) = \max\limits_{y \in Y} \frac{1}{N} \sum\limits_{n=1}^N p^n(y)
\end{equation}

\noindent \emph{Probability Variance} (PV) \cite{gal2016dropout} computes negative variance of them. $\bar{p}_y$ is mean probability for a label $y$ across N values.
\begin{equation}
    \tilde{g}_{PV}(x)= \frac{1}{|Y|}\sum\limits_{y \in Y}\frac{1}{N} \sum\limits_{n=1}^N(p^n_y - \bar{p}_y)^2 
\end{equation}

\noindent \emph{Bayesian active learning by disagreement} (BALD) \cite{houlsby2011bayesian,vazhentsev2022uncertainty}  measures using the mutual information as follows: 
\begin{equation} 
    \tilde{g}_{BALD}(x)=  \sum\limits_{y \in Y} \bar{p}_y  \log \bar{p}_y + \sum\limits_{y,n} p_y^n \log p_y^n 
\end{equation}
This MC dropout mechanism is equivalent to using an ensemble model for confidence estimation, but does not require actually training and storing multiple models but with increased inference cost.

\subsection{Training Loss}
\label{training-loss}
\noindent \textbf{Confident Error Regularizer} \cite{xin2021art} adds an additional regularizer, along with task specific loss, aims to optimize for RPP at training time as if the model's error on example exceeds its error on other example (i.e current example is more difficult), then the confidence on that example should not surpass the confidence on other example. 
\begin{equation}
    L_{CER} =  \sum\limits_{1 \leq i,j \leq N} \Delta_{i,j} \mathds{1}[e_i > e_j]
\end{equation}
\begin{equation}
    \Delta_{i,j} = \max\{0, \max\limits_{y \in Y} p_i(y) - \max\limits_{y \in Y} p_j(y) \}^2
\end{equation}
here N is the number of instances in a batch and $e_i$ is an error of the $i^{th}$ instance and use SR to obtain confidence here as it is easily accessible at training time, while MC-dropout confidence is not.


\begin{figure*}
    \centering
    \scalebox{0.4}{
    \includegraphics{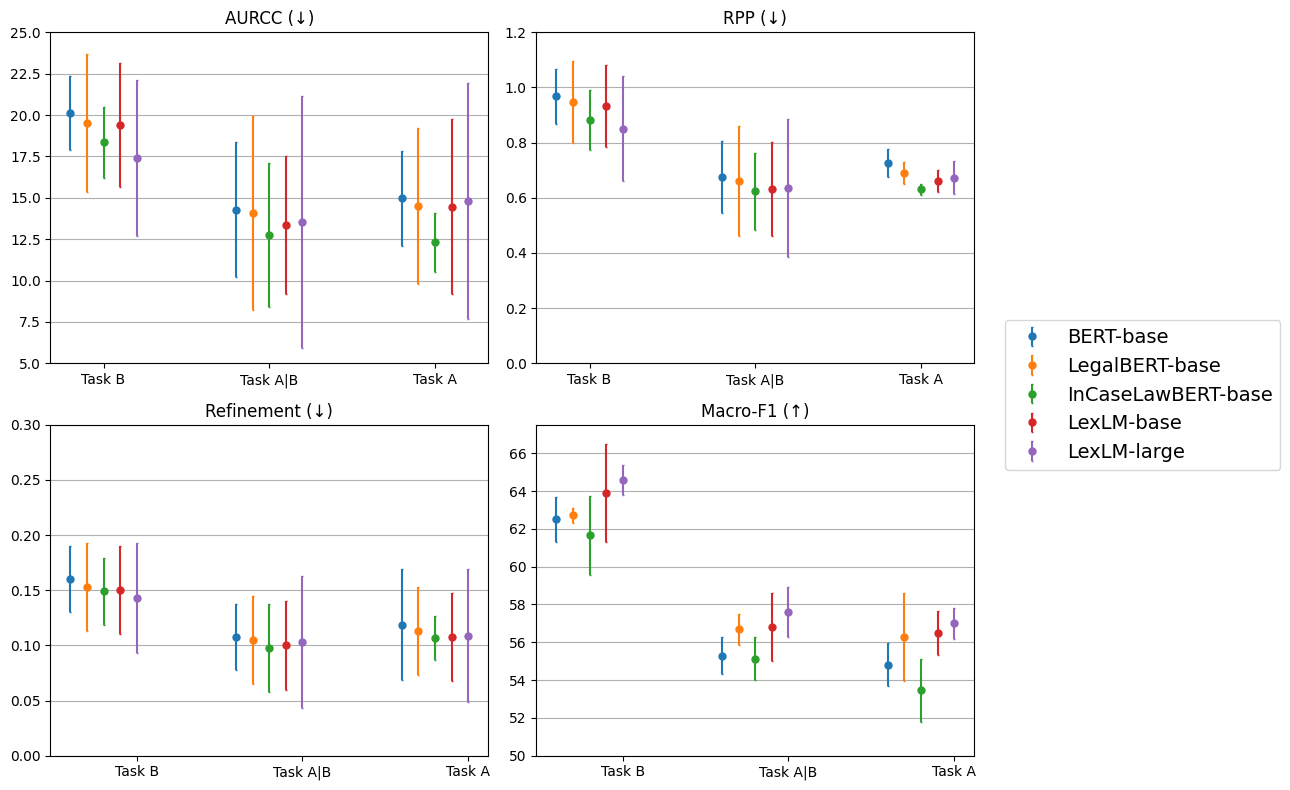}}
    \caption{Impact of pre-trained model on selective prediction and classification performance.}
    \label{fig:model}
\end{figure*}

\noindent \textbf{Expected Calibration Error (ECE) Loss} \cite{shamsiuncertainty} is added additionally to task-specific loss and is calculated by grouping the predictions into different bins (M bins) according to their confidence. The final loss is aggregated across bins where the error of each bin is computed as the difference between the accuracy and the confidence as:

\begin{equation}
\scalemath{0.85}{
L_{ECE} = \sum_{m \in M} \frac{|B_m|}{N} |\text{acc}(B_m) - \text{conf}(B_m)|}
\end{equation}

\begin{equation*}
\scalemath{0.8}{
\text{acc}(B_m) = \frac{1}{|B_m|} \sum_{i \in B_m} \mathds{1}_{\hat{y}_i = y_i}  ~~\&~~ \text{conf}(B_m) = \frac{1}{|B_m|} \sum_{i \in B_m} p_i}
\end{equation*}
\noindent where N represents number of instances, $\hat{y_i}$ and $y_i$ indicate predicted and actual labels respectively.

\noindent \textbf{Gambler`s Loss} \cite{liu2019deep} Unlike above methods which derive abstention label based on confidence estimate of actual label predictions, Gambler's loss explicitly augments an extra class to learn the selection function and trains with the loss function that allows the prediction function to benefit from abstaining on difficult instances:
\begin{equation}
    L_{gambler} = \ \sum\limits_{y \in Y} I(y) \log [{p(y) + \frac{1}{r} {p(abs)]} }
\end{equation}
where $I(y)$ is binary indicator indicating if $y$ is the true label, $p(abs)$ denotes the rejection score and and $r$ is the rejection reward hyperparameter. Here the coverage is obtained by varying the threshold of abstention logit.

\subsection{Extending to Multi-label case}
While all the above techniques are typically proposed for multi-class classification scenarios,  we adapt them to the multi-label setting for our COC task by treating each label as a separate binary classification task. This involves deriving a confidence estimate for each label by utilizing the probability assigned to that label and its complement (1 - probability). Similarly, we vary the threshold for each label independently to facilitate abstention and the evaluation metrics are computed for each label separately. We report the macro-average across all the labels for an instance, unless specified.

\section{Experiments}



\begin{figure*}
    \centering
    \scalebox{0.4}{
    \includegraphics{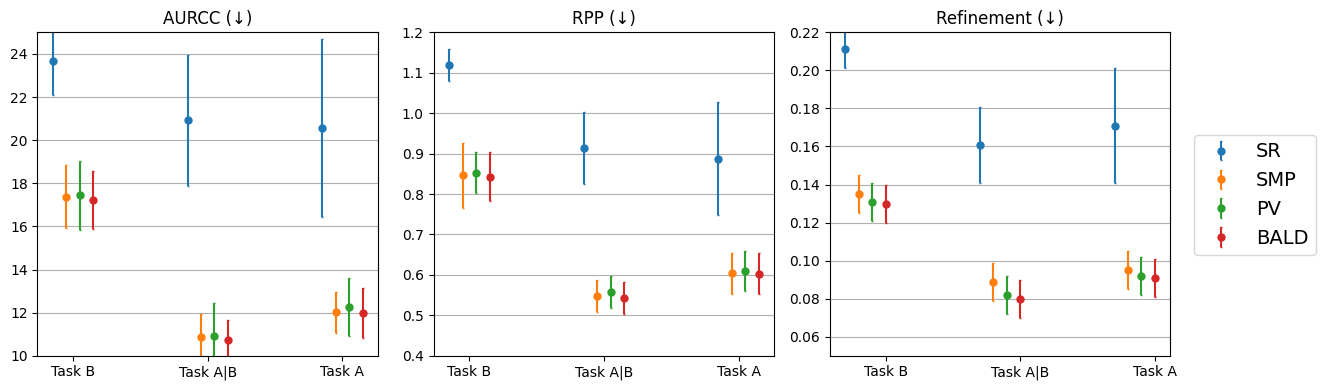}}
    \caption{Impact of confidence estimators on selective prediction and classification performance.}
    \label{fig:conf}
\end{figure*}

Following \citealt{chalkidis2022lexglue}, we employ a hierarchical extension of a pre-trained transformer model to account for longer input texts as our base model. We use corresponding pre-trained model to encode each paragraph in the input independently to obtain [cls] representation for each paragraph. We then pass these paragraph representations to a transformer layer to learn contextual information from other paragraphs. Finally, we max pool over these context-aware paragraph representations to obtain final representation of case facts which is sent to classification layer. In case of Task A|B, we concatenate a multi-hot feature vector containing the task B labels to the final representation before passing it to the classifier as in \citealt{santosh2022deconfounding}. Detailed hyperparameters can be found in App. \ref{impl}.

To assess the reliability of COC models using diverse pre-trained language models as backbone, we employ (i) BERT-base \cite{kenton2019bert}, trained on general english corpora (ii) InCaseLawBERT-base \cite{paul2023pre}, pre-trained on Indian Supreme Court and High Courts case documents by initializing with CaseLawBERT \cite{zheng2021does} which is pre-trained on US Case Law from federal and state courts. (iii) LegalBERT-base \cite{chalkidis2020legal} incorporates pre-training on EU, UK, and cases from the US, European Court of Justice, and ECtHR. (iv) LexLM-base \cite{chalkidis2023lexfiles}, pre-trained on LeXFiles, a diverse legal corpus spanning US, UK, EU, India, and Canada jurisdictions. (v) LexLM-large \cite{chalkidis2023lexfiles}, a larger version pre-trained on the same LexFiles corpus. 
Among them, BERT lacks specific legal pre-training, while InCaseLawBERT has access to legal corpus but not from the ECtHR jurisdiction, relevant to our COC task. LegalBERT and LexLM have access to ECtHR corpus in pre-training, but they differ in the proportion of ECtHR corpus to other corpora, where LexLM has less ECtHR proportion compared to LegalBERT's pre-training corpora.

We use these 5 pre-trained models as backbone in base model and fine-tune on COC task with 4 training loss functions (i.e task-specific loss, CER, ECE, Gambler in Sec. \ref{training-loss}). We employ 4 variants (SR, SMP, PV, BALD in Sec. \ref{confidence}) to derive the confidence estimate and thus compute the selective prediction metrics (AURCC, RPP, Refinement) and classification metric (macro-F1) across these 80 (5*4*4) configurations for each of the three tasks.

\subsection{Results}
\noindent \textbf{(a) Impact of Pre-trained models:} Fig. \ref{fig:model} reports the selective prediction 
and classification performance metrics 
, averaged across all the configurations (confidence estimators, training losses) for each model for three COC tasks along with standard deviation. Lower scores are preferable for selective metrics, while higher scores are desired for macro-F1. Our observations reveal that on macro-F1 scores, BERT consistently outperforms InCaseLawBERT across all three tasks consistently. LegalBERT, LexLM-Base, and LexLM-Large  outperform BERT's performance in that order. We hypothesize that InCaseLawBERT, despite being pre-trained on domain-specific (legal) US and Indian contexts, may not serve as a better starting point for generalizing to the ECtHR COC task. In contrast, models with access to the ECtHR corpus, such as LegalBERT, LexLM-Base, and LexLM-Large, benefit from pre-training, with LexLM's diverse corpus contributing to better generalization capabilities.


\begin{figure*}
    \centering
    \scalebox{0.4}{
    \includegraphics{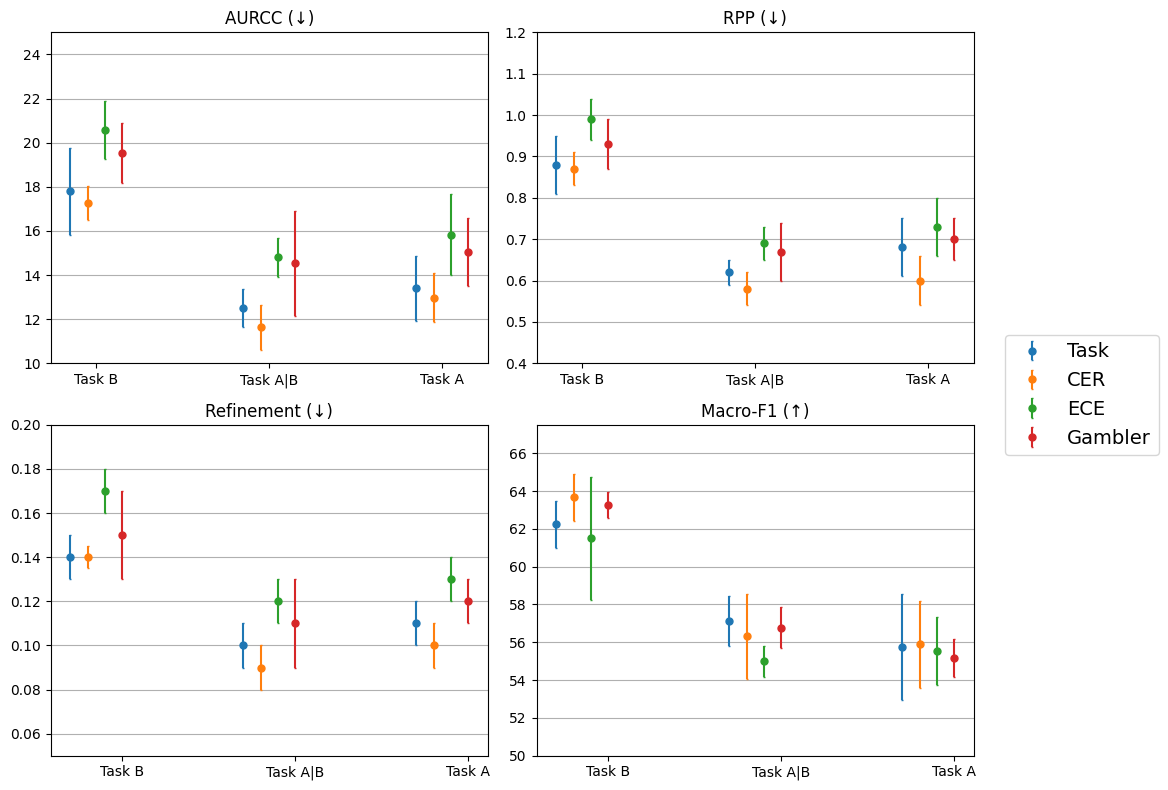}}
    \caption{Impact of training loss function on selective prediction and classification performance.}
    \label{fig:loss}
\end{figure*}

On examining selective metrics, BERT-base exhibits the highest (worst) score compared to legally pre-trained models, underscoring that domain-specific pre-training, can aid not only in better accuracy but also lead to better calibration than general ones. Among the legal base models, the proportion of ECtHR corpus in pre-training inversely correlates with scores on selective metrics—the model with no access (InCaseLawBERT) performs the best, followed by LexLM and LegalBERT. This trend is consistent across the three selective metrics. The tendency of models with access to ECtHR corpus in pre-training resulting in overconfident predictions on downstream tasks, may be attributed to spurious artifacts present in the data, influencing the model right from the pre-training stage. This pattern persists across all three tasks. Surprisingly, LexLM-Large maintains the best selective performance on Task B despite having access to ECtHR in pre-training. This could be attributed to its higher number of parameters enabling better generalization. However, the effect of greater parameterization in LexLM-Large diminishes when transitioning to more challenging tasks (Task A|B,  A), rendering it overconfident. 

\noindent \textbf{Main Takeaways:} Our findings underscore that (i) Access to a domain-specific corpus during pre-training improves model calibration compared to its absence. (ii) However, excessive focus on a downstream domain corpus during pre-training may negatively impact calibration, causing the model to pick up spurious artifacts during pre-training phase that turn challenging to unlearn during fine-tuning. This highlights the need for a diverse domain-related corpus during pre-training to strike a balance between accuracy and confidence. Additionally, the development of effective saliency masking strategies during pre-training is crucial for producing more robust and reliable pre-trained models, steering away from spurious artifacts. (iii) Larger models exhibit overconfidence compared to their base versions, despite achieving higher accuracy.



\begin{figure*}
    \centering
    \scalebox{0.42}{
    \includegraphics{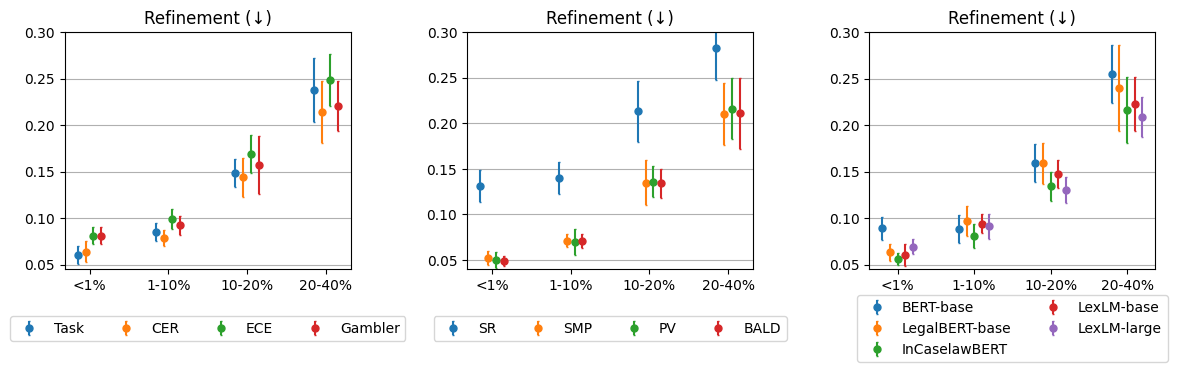}}
    \caption{Impact of pre-trained model, confidence estimator and loss function on different labels bucketed by frequency of cases in which they deem to be violated in Task A.}
    \label{fig:freq}
\end{figure*}

\vspace{0.4em}
\noindent \textbf{(b) Impact of Confidence Estimation} We present averaged selective prediction metrics\footnote{The choice of confidence estimator does not affect the models' performance as it is dependent on the pre-trained model and loss function.} along with standard deviation across various models and training loss function configurations for each confidence estimator in Fig. \ref{fig:conf}. We observe that computationally intensive MC dropout variants - SMP, PV, BALD - achieve significant improvements over SR on all metrics and tasks consistently, aligning with \citealt{vazhentsev2022uncertainty} and contrary to the findings of \citealt{xin2021art}, where the opposite trend is observed. 
\citet{gu2023evaluation} advocate to use refinement to compare confidence estimators, in contrast to AURCC and RPP, commonly used in prior works \cite{xin2021art, vazhentsev2022uncertainty, vazhentsev2023hybrid, whitehead2022reliable} as as they are also influenced by the effectiveness of the base prediction function. On refinement among the three MC methods, BALD takes the slightest lead on all tasks, followed by PV and SMP, albeit marginally. The effectiveness of BALD compared to PV and SMP may stem from the latter focusing exclusively on epistemic uncertainty arising from a lack of knowledge, ignoring aleatoric uncertainty associated with ambiguity and noise in the data, while the former measures total uncertainty \cite{vazhentsev2022uncertainty, malinin2018predictive}.


\noindent \textbf{Main Takeaways:} MC Dropout notably enhances confidence estimation compared to SR, albeit with an increase in computational costs during inference. This raises concerns, considering the substantial computational overhead with these pre-trained models. Therefore, the development of effective yet computationally light confidence estimators represents a potential avenue for further exploration.

\vspace{0.4em}
\noindent \textbf{(c) Impact of Training Loss} We report the selective prediction and performance metrics averaged across all configurations for each training loss in Fig. \ref{fig:loss}. We observe that adding confident error regularizer boosted all the selective metrics consistently across all the tasks than trained with task-specific loss alone. This can be attributed to its design to deliberately alleviate overconfidence during fine-tuning. Notably, it did not harm model accuracy and maintained comparable or even better to baseline in some cases. On the other hand, ECE regularization negatively impacts both confidence calibration and the performance than normal one. This is attributed to the non-differentiable nature of ECE loss, prompting the need for further investigation into differentiable surrogates of ECE loss, as proposed in \cite{karandikar2021soft, bohdal2021meta}. Gamblers loss which allows model to directly learn abstention head, maintained comparable performance to baseline/CER but witnessed a drop in selective prediction performance. 

\noindent \textbf{Main Takeaways:} Adding CER is the most compelling recipe for fine-tuning to balance accuracy with confidence calibration.

\vspace{0.4em}
\noindent \textbf{(d) Effect on different labels} To assess the influence of each design factor on different labels, we categorize the labels based on their frequency of occurrence in cases deemed to be violated in the training set (Task A). Thus we obtain four buckets with 5, 4, 4, 1 articles accounting for <1\% (rarely violated), 1-10\%, 10-20\% and 20-40\% (frequently violated) respectively. We calculate the averaged refinement scores across articles in each bucket, considering various configurations under different training losses, confidence estimators and pre-trained models and present the results in Fig. \ref{fig:freq}. With respect to training loss, we observe that adding CER performs better than baseline task specific loss alone, as we move towards more frequent labels. This can also be due to less number of violated cases available for rare articles to effectively regularize them, as the probability of them appearing in same batch tends to be lower. While ECE and Gamblers underperformed compared to task-specific loss in rare article buckets, but gamblers picked it up towards frequent bucket, making it comparable to CER.

Among confidence estimators, MC dropout methods consistently maintained better performance then SR across all the buckets. 
Across pre-trained models, InCaseLawBERT maintains better performance across all the buckets and trend of decreasing performance with increasing access to ECtHR can also be noticed from LexLM-base and LegalBERT. On the other hand, BERT-base shows the worst performance across all buckets, emphasizing the need of domain-specific pre-training for a better calibrated model. However, LexLM-large suffers in rare article buckets, but picks up in the frequent article buckets due to presence of larger number of parameters to capture diverse signals in frequent cases. Models' overconfidence values increase towards the frequent buckets due to the confounding effect of more positive data instances.

\section{Conclusion}
We introduce the problem of selective prediction for COC task, aiming to enhance model reliability by abstaining in cases of low confidence. Through empirical investigation on 3 COC task variants with 5 pre-trained models, 4 confidence estimators, and 4 loss functions, we assess how these design factors contribute to better selective prediction in COC. Our findings reveal that legal domain-specific pre-trained models outperform in classification-related metrics and are well-calibrated, but an exclusive focus on a specific corpus can prove detrimental. Larger models tend to be overconfident compared to their base versions. Despite being computationally intense, MC Dropout methods provide superior confidence estimates. Adding CER regularization helps alleviate overconfidence. We hope this preliminary investigation spurs additional research into the selective prediction of COC models, emphasizing the critical need for models to be aware of what they don't know in high-stakes domains like Legal COC.

\section*{Limitations}
Our study is limited by the datasets, models and selective prediction techniques we consider. We rely on the ECtHR dataset and the findings may be influenced by any characterstics present in this particular dataset such as spurious correlations in the downstream task and effectiveness of simple token based decision trees for such tasks \cite{santosh2022deconfounding}. Extending the study to diverse datasets and legal jurisdictions would contribute to a more comprehensive understanding of the reliability of design factors associated with Case Outcome Classification models in legal NLP and thus bolstering the generalizability.

Due to computational constraints, we are unable to pre-train language models from scratch with specific combinations, such as exclusive pre-training on the ECtHR corpus, larger model size variants with billions of parameters, or different pre-training corporus combinations, hindering our ability to conduct dedicated ablations on our claims. Consequently, we rely on existing pre-trained models and solely undertake fine-tuning in this study. While we have made efforts to include a diverse range of selective prediction techniques in our empirical investigation, it is important to acknowledge that our study may not comprehensively cover all the work in this space. Moreover, while we have selected certain design factors for examination, there are other architectural variants proposed in recent studies such as dependency learning across tasks, incorporation of loss functions like contrastive learning, integration of additional external legal knowledge, like legal articles. These aspects warrant exploration in future research endeavors to assess their impact on model reliability.

Additionally, we advocate for future studies in COC to not only report prediction performance but also include reliability metrics to provide deeper insights into models' confidence calibration. This would offer a more comprehensive understanding of model behavior and enhance the trustworthiness of COC models in legal contexts.

\section*{Ethics Statement}
Our dataset is derived from LexGLUE benchmark \cite{chalkidis2022lexglue} which is obtained from a publicly available database of ECtHR decisions, available in the public court database HUDOC\footnote{\url{https://hudoc.echr.coe.int}}. Despite the inclusion of real names and the absence of anonymization in these decisions, we do not foresee any harm resulting from our experiments beyond the disclosure of this information.

The task of COC/LJP raises significant ethical and legal considerations, both in a broad context and particularly concerning the European Court of Human Rights \cite{fikfak2021future}. It is crucial to clarify that our research does not advocate for the practical implementation of Case Outcome Classification (COC) within courts. The experimental nature of our study is designed to explore the reliability of models in controlled settings and does not propose or endorse real-world deployment within legal systems. Our results are hence to be understood as technical contributions in pursuit of the overarching goal of developing models capable of deriving insight from data that can be used legally, ethically, and mindfully by experts in solving problems arising in legal research and practice.

We acknowledge that, in adapting pre-trained encoders, our models may inherit existing biases. Similarly, the ECtHR case collection, being historical data, may exhibit a data distribution where sensitive attributes of individuals (e.g., applicant gender) could offer predictive signals for the allegation/violation variable, as demonstrated in previous work \cite{chalkidis2022fairlex}. Although we believe the results observed in our COC experiments are not substantially connected to such encoded bias, it is crucial to highlight that legal NLP systems utilizing case outcome information and intended for practical deployment should undergo thorough scrutiny against relevant equal treatment imperatives, ensuring scrutiny of their performance, behavior, and intended use.

\bibliography{custom}

\begin{thebibliography}{85}
\expandafter\ifx\csname natexlab\endcsname\relax\def\natexlab#1{#1}\fi

\bibitem[{Alali et~al.(2021)Alali, Syed, Alsayed, Patel, and Bodala}]{alali2021justice}
Mohammad Alali, Shaayan Syed, Mohammed Alsayed, Smit Patel, and Hemanth Bodala. 2021.
\newblock Justice: A benchmark dataset for supreme court's judgment prediction.
\newblock \emph{arXiv preprint arXiv:2112.03414}.

\bibitem[{Aletras et~al.(2016)Aletras, Tsarapatsanis, Preo{\c{t}}iuc-Pietro, and Lampos}]{aletras2016predicting}
Nikolaos Aletras, Dimitrios Tsarapatsanis, Daniel Preo{\c{t}}iuc-Pietro, and Vasileios Lampos. 2016.
\newblock Predicting judicial decisions of the european court of human rights: A natural language processing perspective.
\newblock \emph{PeerJ Computer Science}, 2:e93.

\bibitem[{Baan et~al.(2024)Baan, Fern{\'a}ndez, Plank, and Aziz}]{baan2024interpreting}
Joris Baan, Raquel Fern{\'a}ndez, Barbara Plank, and Wilker Aziz. 2024.
\newblock Interpreting predictive probabilities: Model confidence or human label variation?
\newblock \emph{arXiv preprint arXiv:2402.16102}.

\bibitem[{Baumgartner et~al.(2024)Baumgartner, St{\"u}rmer, Grabmair, Niklaus et~al.}]{baumgartner2024towards}
Nina Baumgartner, Matthias St{\"u}rmer, Matthias Grabmair, Joel Niklaus, et~al. 2024.
\newblock Towards explainability and fairness in swiss judgement prediction: Benchmarking on a multilingual dataset.
\newblock \emph{arXiv preprint arXiv:2402.17013}.

\bibitem[{Bohdal et~al.(2021)Bohdal, Yang, and Hospedales}]{bohdal2021meta}
Ondrej Bohdal, Yongxin Yang, and Timothy Hospedales. 2021.
\newblock Meta-calibration: Learning of model calibration using differentiable expected calibration error.
\newblock \emph{arXiv preprint arXiv:2106.09613}.

\bibitem[{Chalkidis et~al.(2019)Chalkidis, Androutsopoulos, and Aletras}]{chalkidis2019neural}
Ilias Chalkidis, Ion Androutsopoulos, and Nikolaos Aletras. 2019.
\newblock Neural legal judgment prediction in english.
\newblock In \emph{Proceedings of the 57th Annual Meeting of the Association for Computational Linguistics}, pages 4317--4323.

\bibitem[{Chalkidis et~al.(2020)Chalkidis, Fergadiotis, Malakasiotis, Aletras, and Androutsopoulos}]{chalkidis2020legal}
Ilias Chalkidis, Manos Fergadiotis, Prodromos Malakasiotis, Nikolaos Aletras, and Ion Androutsopoulos. 2020.
\newblock Legal-bert: The muppets straight out of law school.
\newblock In \emph{Findings of the Association for Computational Linguistics: EMNLP 2020}, pages 2898--2904.

\bibitem[{Chalkidis et~al.(2021)Chalkidis, Fergadiotis, Tsarapatsanis, Aletras, Androutsopoulos, and Malakasiotis}]{chalkidis2021paragraph}
Ilias Chalkidis, Manos Fergadiotis, Dimitrios Tsarapatsanis, Nikolaos Aletras, Ion Androutsopoulos, and Prodromos Malakasiotis. 2021.
\newblock Paragraph-level rationale extraction through regularization: A case study on european court of human rights cases.
\newblock \emph{arXiv preprint arXiv:2103.13084}.

\bibitem[{Chalkidis et~al.(2023)Chalkidis, Garneau, Goanta, Katz, and S{\o}gaard}]{chalkidis2023lexfiles}
Ilias Chalkidis, Nicolas Garneau, Catalina Goanta, Daniel~Martin Katz, and Anders S{\o}gaard. 2023.
\newblock Lexfiles and legallama: Facilitating english multinational legal language model development.
\newblock \emph{arXiv preprint arXiv:2305.07507}.

\bibitem[{Chalkidis et~al.(2022{\natexlab{a}})Chalkidis, Jana, Hartung, Bommarito, Androutsopoulos, Katz, and Aletras}]{chalkidis2022lexglue}
Ilias Chalkidis, Abhik Jana, Dirk Hartung, Michael Bommarito, Ion Androutsopoulos, Daniel Katz, and Nikolaos Aletras. 2022{\natexlab{a}}.
\newblock Lexglue: A benchmark dataset for legal language understanding in english.
\newblock In \emph{Proceedings of the 60th Annual Meeting of the Association for Computational Linguistics (Volume 1: Long Papers)}, pages 4310--4330.

\bibitem[{Chalkidis et~al.(2022{\natexlab{b}})Chalkidis, Pasini, Zhang, Tomada, Schwemer, and S{\o}gaard}]{chalkidis2022fairlex}
Ilias Chalkidis, Tommaso Pasini, Sheng Zhang, Letizia Tomada, Sebastian~Felix Schwemer, and Anders S{\o}gaard. 2022{\natexlab{b}}.
\newblock Fairlex: A multilingual benchmark for evaluating fairness in legal text processing.
\newblock \emph{arXiv preprint arXiv:2203.07228}.

\bibitem[{Chen et~al.(2023)Chen, Yoon, Ebrahimi, Arik, Pfister, and Jha}]{chen2023adaptation}
Jiefeng Chen, Jinsung Yoon, Sayna Ebrahimi, Sercan~O Arik, Tomas Pfister, and Somesh Jha. 2023.
\newblock Adaptation with self-evaluation to improve selective prediction in llms.
\newblock \emph{arXiv preprint arXiv:2310.11689}.

\bibitem[{Chow(1957)}]{chow1957optimum}
Chi-Keung Chow. 1957.
\newblock An optimum character recognition system using decision functions.
\newblock \emph{IRE Transactions on Electronic Computers}, (4):247--254.

\bibitem[{Cole et~al.(2023)Cole, Zhang, Gillick, Eisenschlos, Dhingra, and Eisenstein}]{cole2023selectively}
Jeremy~R Cole, Michael~JQ Zhang, Daniel Gillick, Julian~Martin Eisenschlos, Bhuwan Dhingra, and Jacob Eisenstein. 2023.
\newblock Selectively answering ambiguous questions.
\newblock \emph{arXiv preprint arXiv:2305.14613}.

\bibitem[{Cortes et~al.(2016)Cortes, DeSalvo, and Mohri}]{cortes2016learning}
Corinna Cortes, Giulia DeSalvo, and Mehryar Mohri. 2016.
\newblock Learning with rejection.
\newblock In \emph{Algorithmic Learning Theory: 27th International Conference, ALT 2016, Bari, Italy, October 19-21, 2016, Proceedings 27}, pages 67--82. Springer.

\bibitem[{Desai and Durrett(2020)}]{desai2020calibration}
Shrey Desai and Greg Durrett. 2020.
\newblock Calibration of pre-trained transformers.
\newblock In \emph{Proceedings of the 2020 Conference on Empirical Methods in Natural Language Processing (EMNLP)}, pages 295--302.

\bibitem[{Douka et~al.(2021)Douka, Abdine, Vazirgiannis, El~Hamdani, and Amariles}]{douka2021juribert}
Stella Douka, Hadi Abdine, Michalis Vazirgiannis, Rajaa El~Hamdani, and David~Restrepo Amariles. 2021.
\newblock Juribert: A masked-language model adaptation for french legal text.
\newblock In \emph{Proceedings of the Natural Legal Language Processing Workshop 2021}, pages 95--101.

\bibitem[{El-Yaniv et~al.(2010)}]{el2010foundations}
Ran El-Yaniv et~al. 2010.
\newblock On the foundations of noise-free selective classification.
\newblock \emph{Journal of Machine Learning Research}, 11(5).

\bibitem[{Fikfak(2021)}]{fikfak2021future}
Veronika Fikfak. 2021.
\newblock What future for human rights? decision-making by algorithm.
\newblock \emph{Decision-making by algorithm (September 3, 2021). Strasbourg Observers}, 19.

\bibitem[{Fumera and Roli(2002)}]{fumera2002support}
Giorgio Fumera and Fabio Roli. 2002.
\newblock Support vector machines with embedded reject option.
\newblock In \emph{Pattern Recognition with Support Vector Machines: First International Workshop, SVM 2002 Niagara Falls, Canada, August 10, 2002 Proceedings}, pages 68--82. Springer.

\bibitem[{Gal and Ghahramani(2016)}]{gal2016dropout}
Yarin Gal and Zoubin Ghahramani. 2016.
\newblock Dropout as a bayesian approximation: Representing model uncertainty in deep learning.
\newblock In \emph{international conference on machine learning}, pages 1050--1059. PMLR.

\bibitem[{Gan et~al.(2021)Gan, Kuang, Yang, and Wu}]{gan2021judgment}
Leilei Gan, Kun Kuang, Yi~Yang, and Fei Wu. 2021.
\newblock Judgment prediction via injecting legal knowledge into neural networks.
\newblock In \emph{Proceedings of the AAAI Conference on Artificial Intelligence}, volume~35, pages 12866--12874.

\bibitem[{Garg and Moschitti(2021)}]{garg2021will}
Siddhant Garg and Alessandro Moschitti. 2021.
\newblock Will this question be answered? question filtering via answer model distillation for efficient question answering.
\newblock In \emph{Proceedings of the 2021 Conference on Empirical Methods in Natural Language Processing}, pages 7329--7346.

\bibitem[{Geifman and El-Yaniv(2017)}]{geifman2017selective}
Yonatan Geifman and Ran El-Yaniv. 2017.
\newblock Selective classification for deep neural networks.
\newblock \emph{Advances in neural information processing systems}, 30.

\bibitem[{Gu and Hopkins(2023)}]{gu2023evaluation}
Zhengyao Gu and Mark Hopkins. 2023.
\newblock On the evaluation of neural selective prediction methods for natural language processing.
\newblock In \emph{Proceedings of the 61st Annual Meeting of the Association for Computational Linguistics (Volume 1: Long Papers)}, pages 7888--7899.

\bibitem[{Guo et~al.(2017)Guo, Pleiss, Sun, and Weinberger}]{guo2017calibration}
Chuan Guo, Geoff Pleiss, Yu~Sun, and Kilian~Q Weinberger. 2017.
\newblock On calibration of modern neural networks.
\newblock In \emph{International conference on machine learning}, pages 1321--1330. PMLR.

\bibitem[{Hellman(1970)}]{hellman1970nearest}
Martin~E Hellman. 1970.
\newblock The nearest neighbor classification rule with a reject option.
\newblock \emph{IEEE Transactions on Systems Science and Cybernetics}, 6(3):179--185.

\bibitem[{Hendrycks and Gimpel(2016)}]{hendrycks2016baseline}
Dan Hendrycks and Kevin Gimpel. 2016.
\newblock A baseline for detecting misclassified and out-of-distribution examples in neural networks.
\newblock In \emph{International Conference on Learning Representations}.

\bibitem[{Houlsby et~al.(2011)Houlsby, Husz{\'a}r, Ghahramani, and Lengyel}]{houlsby2011bayesian}
Neil Houlsby, Ferenc Husz{\'a}r, Zoubin Ghahramani, and M{\'a}t{\'e} Lengyel. 2011.
\newblock Bayesian active learning for classification and preference learning.
\newblock \emph{stat}, 1050:24.

\bibitem[{Hwang et~al.(2022)Hwang, Lee, Cho, Lee, and Seo}]{hwang2022multi}
Wonseok Hwang, Dongjun Lee, Kyoungyeon Cho, Hanuhl Lee, and Minjoon Seo. 2022.
\newblock A multi-task benchmark for korean legal language understanding and judgement prediction.
\newblock \emph{Advances in Neural Information Processing Systems}, 35:32537--32551.

\bibitem[{Jiang et~al.(2018)Jiang, Kim, Guan, and Gupta}]{jiang2018trust}
Heinrich Jiang, Been Kim, Melody Guan, and Maya Gupta. 2018.
\newblock To trust or not to trust a classifier.
\newblock \emph{Advances in neural information processing systems}, 31.

\bibitem[{Kamath et~al.(2020)Kamath, Jia, and Liang}]{kamath2020selective}
Amita Kamath, Robin Jia, and Percy Liang. 2020.
\newblock Selective question answering under domain shift.
\newblock In \emph{Proceedings of the 58th Annual Meeting of the Association for Computational Linguistics}, pages 5684--5696.

\bibitem[{Karandikar et~al.(2021)Karandikar, Cain, Tran, Lakshminarayanan, Shlens, Mozer, and Roelofs}]{karandikar2021soft}
Archit Karandikar, Nicholas Cain, Dustin Tran, Balaji Lakshminarayanan, Jonathon Shlens, Michael~C Mozer, and Becca Roelofs. 2021.
\newblock Soft calibration objectives for neural networks.
\newblock \emph{Advances in Neural Information Processing Systems}, 34:29768--29779.

\bibitem[{Katz et~al.(2017)Katz, Bommarito, and Blackman}]{katz2017general}
Daniel~Martin Katz, Michael~J Bommarito, and Josh Blackman. 2017.
\newblock A general approach for predicting the behavior of the supreme court of the united states.
\newblock \emph{PloS one}, 12(4):e0174698.

\bibitem[{Kaufman et~al.(2019)Kaufman, Kraft, and Sen}]{kaufman2019improving}
Aaron~Russell Kaufman, Peter Kraft, and Maya Sen. 2019.
\newblock Improving supreme court forecasting using boosted decision trees.
\newblock \emph{Political Analysis}, 27(3):381--387.

\bibitem[{Kendall(1948)}]{kendall1948rank}
Maurice~George Kendall. 1948.
\newblock Rank correlation methods.

\bibitem[{Kenton and Toutanova(2019)}]{kenton2019bert}
Jacob Devlin Ming-Wei~Chang Kenton and Lee~Kristina Toutanova. 2019.
\newblock Bert: Pre-training of deep bidirectional transformers for language understanding.
\newblock In \emph{Proceedings of naacL-HLT}, volume~1, page~2.

\bibitem[{Kingma and Ba(2014)}]{kingma2014adam}
Diederik~P Kingma and Jimmy Ba. 2014.
\newblock Adam: A method for stochastic optimization.
\newblock \emph{arXiv preprint arXiv:1412.6980}.

\bibitem[{Kort(1957)}]{kort1957predicting}
Fred Kort. 1957.
\newblock Predicting supreme court decisions mathematically: A quantitative analysis of the “right to counsel” cases.
\newblock \emph{American Political Science Review}, 51(1):1--12.

\bibitem[{Kowsrihawat et~al.(2018)Kowsrihawat, Vateekul, and Boonkwan}]{kowsrihawat2018predicting}
Kankawin Kowsrihawat, Peerapon Vateekul, and Prachya Boonkwan. 2018.
\newblock Predicting judicial decisions of criminal cases from thai supreme court using bi-directional gru with attention mechanism.
\newblock In \emph{2018 5th Asian Conference on Defense Technology (ACDT)}, pages 50--55. IEEE.

\bibitem[{Lage-Freitas et~al.(2022)Lage-Freitas, Allende-Cid, Santana, and Oliveira-Lage}]{lage2022predicting}
Andr{\'e} Lage-Freitas, H{\'e}ctor Allende-Cid, Orivaldo Santana, and L{\'\i}via Oliveira-Lage. 2022.
\newblock Predicting brazilian court decisions.
\newblock \emph{PeerJ Computer Science}, 8:e904.

\bibitem[{Li et~al.(2022)Li, Huang, Geng, Guo, and Yuan}]{li2022fairness}
Yanjun Li, Huan Huang, Qiang Geng, Xinwei Guo, and Yuyu Yuan. 2022.
\newblock Fairness measures of machine learning models in judicial penalty prediction.
\newblock \emph{Journal of Internet Technology}, 23(5):1109--1116.

\bibitem[{Liu et~al.(2023)Liu, Wu, Zhang, Sun, Lu, Wu, and Kuang}]{liu2023ml}
Yifei Liu, Yiquan Wu, Yating Zhang, Changlong Sun, Weiming Lu, Fei Wu, and Kun Kuang. 2023.
\newblock Ml-ljp: Multi-law aware legal judgment prediction.
\newblock In \emph{Proceedings of the 46th International ACM SIGIR Conference on Research and Development in Information Retrieval}, pages 1023--1034.

\bibitem[{Liu et~al.(2019)Liu, Wang, Liang, Salakhutdinov, Morency, and Ueda}]{liu2019deep}
Ziyin Liu, Zhikang Wang, Paul~Pu Liang, Russ~R Salakhutdinov, Louis-Philippe Morency, and Masahito Ueda. 2019.
\newblock Deep gamblers: Learning to abstain with portfolio theory.
\newblock \emph{Advances in Neural Information Processing Systems}, 32.

\bibitem[{Malik et~al.(2021)Malik, Sanjay, Nigam, Ghosh, Guha, Bhattacharya, and Modi}]{malik2021ildc}
Vijit Malik, Rishabh Sanjay, Shubham~Kumar Nigam, Kripabandhu Ghosh, Shouvik~Kumar Guha, Arnab Bhattacharya, and Ashutosh Modi. 2021.
\newblock Ildc for cjpe: Indian legal documents corpus for court judgment prediction and explanation.
\newblock In \emph{Proceedings of the 59th Annual Meeting of the Association for Computational Linguistics and the 11th International Joint Conference on Natural Language Processing (Volume 1: Long Papers)}, pages 4046--4062.

\bibitem[{Malinin and Gales(2018)}]{malinin2018predictive}
Andrey Malinin and Mark Gales. 2018.
\newblock Predictive uncertainty estimation via prior networks.
\newblock \emph{Advances in neural information processing systems}, 31.

\bibitem[{Medvedeva et~al.(2021)Medvedeva, {\"U}st{\"u}n, Xu, Vols, and Wieling}]{medvedeva2021automatic}
Masha Medvedeva, Ahmet {\"U}st{\"u}n, Xiao Xu, Michel Vols, and Martijn Wieling. 2021.
\newblock Automatic judgement forecasting for pending applications of the european court of human rights.
\newblock In \emph{ASAIL/LegalAIIA@ ICAIL}, pages 12--23.

\bibitem[{Nagel(1963)}]{nagel1963applying}
Stuart~S Nagel. 1963.
\newblock Applying correlation analysis to case prediction.
\newblock \emph{Tex. L. Rev.}, 42:1006.

\bibitem[{Niklaus et~al.(2021)Niklaus, Chalkidis, and St{\"u}rmer}]{niklaus2021swiss}
Joel Niklaus, Ilias Chalkidis, and Matthias St{\"u}rmer. 2021.
\newblock Swiss-judgment-prediction: A multilingual legal judgment prediction benchmark.
\newblock In \emph{Proceedings of the Natural Legal Language Processing Workshop 2021}, pages 19--35.

\bibitem[{Paul et~al.(2023)Paul, Mandal, Goyal, and Ghosh}]{paul2023pre}
Shounak Paul, Arpan Mandal, Pawan Goyal, and Saptarshi Ghosh. 2023.
\newblock Pre-trained language models for the legal domain: a case study on indian law.
\newblock In \emph{Proceedings of the Nineteenth International Conference on Artificial Intelligence and Law}, pages 187--196.

\bibitem[{Ren et~al.(2022)Ren, Luo, Zhao, Krishna, Saleh, Lakshminarayanan, and Liu}]{ren2022out}
Jie Ren, Jiaming Luo, Yao Zhao, Kundan Krishna, Mohammad Saleh, Balaji Lakshminarayanan, and Peter~J Liu. 2022.
\newblock Out-of-distribution detection and selective generation for conditional language models.
\newblock In \emph{The Eleventh International Conference on Learning Representations}.

\bibitem[{Santosh et~al.(2023{\natexlab{a}})Santosh, Ichim, and Grabmair}]{santosh2023Zero}
T.~Y. S.~S Santosh, Oana Ichim, and Matthias Grabmair. 2023{\natexlab{a}}.
\newblock Zero shot transfer of article-aware legal outcome classification for european court of human rights cases.
\newblock \emph{arXiv preprint arXiv:2302.00609}.

\bibitem[{Santosh et~al.(2023{\natexlab{b}})Santosh, Blas, Kemper, and Grabmair}]{santosh2023leveraging}
TYS Santosh, Marcel Perez~San Blas, Phillip Kemper, and Matthias Grabmair. 2023{\natexlab{b}}.
\newblock Leveraging task dependency and contrastive learning for case outcome classification on european court of human rights cases.
\newblock \emph{arXiv preprint arXiv:2302.00768}.

\bibitem[{Santosh et~al.(2022{\natexlab{a}})Santosh, Xu, Ichim, and Grabmair}]{santosh2022deconfounding}
TYS Santosh, Shanshan Xu, Oana Ichim, and Matthias Grabmair. 2022{\natexlab{a}}.
\newblock Deconfounding legal judgment prediction for european court of human rights cases towards better alignment with experts.
\newblock \emph{arXiv preprint arXiv:2210.13836}.

\bibitem[{Santosh et~al.(2022{\natexlab{b}})Santosh, Xu, Ichim, and Grabmair}]{santosh2022deconfounding2}
T.y.s.s Santosh, Shanshan Xu, Oana Ichim, and Matthias Grabmair. 2022{\natexlab{b}}.
\newblock \href {https://aclanthology.org/2022.emnlp-main.74} {Deconfounding legal judgment prediction for {E}uropean court of human rights cases towards better alignment with experts}.
\newblock In \emph{Proceedings of the 2022 Conference on Empirical Methods in Natural Language Processing}, pages 1120--1138, Abu Dhabi, United Arab Emirates. Association for Computational Linguistics.

\bibitem[{Segal(1984)}]{segal1984predicting}
Jeffrey~A Segal. 1984.
\newblock Predicting supreme court cases probabilistically: The search and seizure cases, 1962-1981.
\newblock \emph{American Political Science Review}, 78(4):891--900.

\bibitem[{Shaikh et~al.(2020)Shaikh, Sahu, and Anand}]{shaikh2020predicting}
Rafe~Athar Shaikh, Tirath~Prasad Sahu, and Veena Anand. 2020.
\newblock Predicting outcomes of legal cases based on legal factors using classifiers.
\newblock \emph{Procedia Computer Science}, 167:2393--2402.

\bibitem[{Shamsi et~al.()Shamsi, Asgharnezhad, Tajally, Nahavandi, and Leung}]{shamsiuncertainty}
Afshar Shamsi, Hamzeh Asgharnezhad, AmirReza Tajally, Saeid Nahavandi, and Henry Leung.
\newblock An uncertainty-aware loss function for training neural networks with calibrated predictions.

\bibitem[{Strickson and De~La~Iglesia(2020)}]{strickson2020legal}
Benjamin Strickson and Beatriz De~La~Iglesia. 2020.
\newblock Legal judgement prediction for uk courts.
\newblock In \emph{Proceedings of the 3rd International Conference on Information Science and Systems}, pages 204--209.

\bibitem[{{\c{S}}ulea et~al.(2017{\natexlab{a}}){\c{S}}ulea, Zampieri, Malmasi, Vela, Dinu, and van Genabith}]{csulea2017exploring}
Octavia-Maria {\c{S}}ulea, Marcos Zampieri, Shervin Malmasi, Mihaela Vela, Liviu~P Dinu, and Josef van Genabith. 2017{\natexlab{a}}.
\newblock Exploring the use of text classification in the legal domain.

\bibitem[{{\c{S}}ulea et~al.(2017{\natexlab{b}}){\c{S}}ulea, Zampieri, Vela, and van Genabith}]{csulea2017predicting}
Octavia-Maria {\c{S}}ulea, Marcos Zampieri, Mihaela Vela, and Josef van Genabith. 2017{\natexlab{b}}.
\newblock Predicting the law area and decisions of french supreme court cases.
\newblock In \emph{Proceedings of the International Conference Recent Advances in Natural Language Processing, RANLP 2017}, pages 716--722.

\bibitem[{Tyss et~al.(2023{\natexlab{a}})Tyss, Ichim, and Grabmair}]{tyss2023zero}
Santosh Tyss, Oana Ichim, and Matthias Grabmair. 2023{\natexlab{a}}.
\newblock Zero-shot transfer of article-aware legal outcome classification for european court of human rights cases.
\newblock In \emph{Findings of the Association for Computational Linguistics: EACL 2023}, pages 593--605.

\bibitem[{Tyss et~al.(2023{\natexlab{b}})Tyss, San~Blas, Kemper, and Grabmair}]{tyss2023leveraging}
Santosh Tyss, Marcel~Perez San~Blas, Phillip Kemper, and Matthias Grabmair. 2023{\natexlab{b}}.
\newblock Leveraging task dependency and contrastive learning for case outcome classification on european court of human rights cases.
\newblock In \emph{Proceedings of the 17th Conference of the European Chapter of the Association for Computational Linguistics}, pages 1103--1103.

\bibitem[{Valvoda et~al.(2023)Valvoda, Cotterell, and Teufel}]{valvoda2023role}
Josef Valvoda, Ryan Cotterell, and Simone Teufel. 2023.
\newblock On the role of negative precedent in legal outcome prediction.
\newblock \emph{Transactions of the Association for Computational Linguistics}, 11:34--48.

\bibitem[{Varshney et~al.(2022{\natexlab{a}})Varshney, Mishra, and Baral}]{varshney2022investigating}
Neeraj Varshney, Swaroop Mishra, and Chitta Baral. 2022{\natexlab{a}}.
\newblock Investigating selective prediction approaches across several tasks in iid, ood, and adversarial settings.
\newblock In \emph{60th Annual Meeting of the Association for Computational Linguistics, ACL 2022}, pages 1995--2002. Association for Computational Linguistics (ACL).

\bibitem[{Varshney et~al.(2022{\natexlab{b}})Varshney, Mishra, and Baral}]{varshney2022towards}
Neeraj Varshney, Swaroop Mishra, and Chitta Baral. 2022{\natexlab{b}}.
\newblock Towards improving selective prediction ability of nlp systems.
\newblock In \emph{Proceedings of the 7th Workshop on Representation Learning for NLP}, pages 221--226.

\bibitem[{Vazhentsev et~al.(2022)Vazhentsev, Kuzmin, Shelmanov, Tsvigun, Tsymbalov, Fedyanin, Panov, Panchenko, Gusev, Burtsev et~al.}]{vazhentsev2022uncertainty}
Artem Vazhentsev, Gleb Kuzmin, Artem Shelmanov, Akim Tsvigun, Evgenii Tsymbalov, Kirill Fedyanin, Maxim Panov, Alexander Panchenko, Gleb Gusev, Mikhail Burtsev, et~al. 2022.
\newblock Uncertainty estimation of transformer predictions for misclassification detection.
\newblock In \emph{Proceedings of the 60th Annual Meeting of the Association for Computational Linguistics (Volume 1: Long Papers)}, pages 8237--8252.

\bibitem[{Vazhentsev et~al.(2023)Vazhentsev, Kuzmin, Tsvigun, Panchenko, Panov, Burtsev, and Shelmanov}]{vazhentsev2023hybrid}
Artem Vazhentsev, Gleb Kuzmin, Akim Tsvigun, Alexander Panchenko, Maxim Panov, Mikhail Burtsev, and Artem Shelmanov. 2023.
\newblock Hybrid uncertainty quantification for selective text classification in ambiguous tasks.
\newblock In \emph{Proceedings of the 61st Annual Meeting of the Association for Computational Linguistics (Volume 1: Long Papers)}, pages 11659--11681.

\bibitem[{Virtucio et~al.(2018)Virtucio, Aborot, Abonita, Avinante, Copino, Neverida, Osiana, Peramo, Syjuco, and Tan}]{virtucio2018predicting}
Michael Benedict~L Virtucio, Jeffrey~A Aborot, John Kevin~C Abonita, Roxanne~S Avinante, Rother Jay~B Copino, Michelle~P Neverida, Vanesa~O Osiana, Elmer~C Peramo, Joanna~G Syjuco, and Glenn Brian~A Tan. 2018.
\newblock Predicting decisions of the philippine supreme court using natural language processing and machine learning.
\newblock In \emph{2018 IEEE 42nd annual computer software and applications conference (COMPSAC)}, volume~2, pages 130--135. IEEE.

\bibitem[{Waltl et~al.(2017)Waltl, Bonczek, Scepankova, Landthaler, and Matthes}]{waltl2017predicting}
Bernhard Waltl, Georg Bonczek, Elena Scepankova, J{\"o}rg Landthaler, and Florian Matthes. 2017.
\newblock Predicting the outcome of appeal decisions in germany’s tax law.
\newblock In \emph{International conference on electronic participation}, pages 89--99. Springer.

\bibitem[{Wang et~al.(2020)Wang, Tu, Shi, and Liu}]{wang2020inference}
Shuo Wang, Zhaopeng Tu, Shuming Shi, and Yang Liu. 2020.
\newblock On the inference calibration of neural machine translation.
\newblock \emph{arXiv preprint arXiv:2005.00963}.

\bibitem[{Wang et~al.(2021)Wang, Xiao, Ma, Zhong, Tu, Zhang, Liu, and Sun}]{wang2021equality}
Yuzhong Wang, Chaojun Xiao, Shirong Ma, Haoxi Zhong, Cunchao Tu, Tianyang Zhang, Zhiyuan Liu, and Maosong Sun. 2021.
\newblock Equality before the law: Legal judgment consistency analysis for fairness.
\newblock \emph{arXiv preprint arXiv:2103.13868}.

\bibitem[{Whitehead et~al.(2022)Whitehead, Petryk, Shakib, Gonzalez, Darrell, Rohrbach, and Rohrbach}]{whitehead2022reliable}
Spencer Whitehead, Suzanne Petryk, Vedaad Shakib, Joseph Gonzalez, Trevor Darrell, Anna Rohrbach, and Marcus Rohrbach. 2022.
\newblock Reliable visual question answering: Abstain rather than answer incorrectly.
\newblock In \emph{European Conference on Computer Vision}, pages 148--166. Springer.

\bibitem[{Xiao et~al.(2021)Xiao, Hu, Liu, Tu, and Sun}]{xiao2021lawformer}
Chaojun Xiao, Xueyu Hu, Zhiyuan Liu, Cunchao Tu, and Maosong Sun. 2021.
\newblock Lawformer: A pre-trained language model for chinese legal long documents.
\newblock \emph{AI Open}, 2:79--84.

\bibitem[{Xin et~al.(2021)Xin, Tang, Yu, and Lin}]{xin2021art}
Ji~Xin, Raphael Tang, Yaoliang Yu, and Jimmy Lin. 2021.
\newblock The art of abstention: Selective prediction and error regularization for natural language processing.
\newblock In \emph{Proceedings of the 59th Annual Meeting of the Association for Computational Linguistics and the 11th International Joint Conference on Natural Language Processing (Volume 1: Long Papers)}, pages 1040--1051.

\bibitem[{Xu et~al.(2023)Xu, Ichim, Risini, Plank, Grabmair et~al.}]{xu2023dissonance}
Shanshan Xu, Oana Ichim, Isabella Risini, Barbara Plank, Matthias Grabmair, et~al. 2023.
\newblock From dissonance to insights: Dissecting disagreements in rationale construction for case outcome classification.
\newblock \emph{arXiv preprint arXiv:2310.11878}.

\bibitem[{Xu et~al.(2024)Xu, Santosh, Ichim, Plank, and Grabmair}]{xu2024through}
Shanshan Xu, TYS Santosh, Oana Ichim, Barbara Plank, and Matthias Grabmair. 2024.
\newblock Through the lens of split vote: Exploring disagreement, difficulty and calibration in legal case outcome classification.
\newblock \emph{arXiv preprint arXiv:2402.07214}.

\bibitem[{Yang et~al.(2019)Yang, Jia, Zhou, and Luo}]{yang2019legal}
Wenmian Yang, Weijia Jia, Xiaojie Zhou, and Yutao Luo. 2019.
\newblock Legal judgment prediction via multi-perspective bi-feedback network.
\newblock In \emph{Proceedings of the 28th International Joint Conference on Artificial Intelligence}, pages 4085--4091.

\bibitem[{Yoshikawa and Okazaki(2023)}]{yoshikawa2023selective}
Hiyori Yoshikawa and Naoaki Okazaki. 2023.
\newblock Selective-lama: Selective prediction for confidence-aware evaluation of language models.
\newblock In \emph{Findings of the Association for Computational Linguistics: EACL 2023}, pages 1972--1983.

\bibitem[{Yue et~al.(2021)Yue, Liu, Jin, Wu, Zhang, An, Cheng, Yin, and Wu}]{yue2021neurjudge}
Linan Yue, Qi~Liu, Binbin Jin, Han Wu, Kai Zhang, Yanqing An, Mingyue Cheng, Biao Yin, and Dayong Wu. 2021.
\newblock Neurjudge: A circumstance-aware neural framework for legal judgment prediction.
\newblock In \emph{Proceedings of the 44th International ACM SIGIR Conference on Research and Development in Information Retrieval}, pages 973--982.

\bibitem[{Zhang et~al.(2023)Zhang, Dou, Zhu, and Wen}]{zhang2023contrastive}
Han Zhang, Zhicheng Dou, Yutao Zhu, and Ji-Rong Wen. 2023.
\newblock Contrastive learning for legal judgment prediction.
\newblock \emph{ACM Transactions on Information Systems}, 41(4):1--25.

\bibitem[{Zhao et~al.(2022)Zhao, Yue, An, Zhang, Yu, Liu, and Chen}]{zhao2022cpee}
Lili Zhao, Linan Yue, Yanqing An, Yuren Zhang, Jun Yu, Qi~Liu, and Enhong Chen. 2022.
\newblock Cpee: C ivil case judgment p rediction centering on the trial mode of e ssential e lements.
\newblock In \emph{Proceedings of the 31st ACM International Conference on Information \& Knowledge Management}, pages 2691--2700.

\bibitem[{Zheng et~al.(2021)Zheng, Guha, Anderson, Henderson, and Ho}]{zheng2021does}
Lucia Zheng, Neel Guha, Brandon~R Anderson, Peter Henderson, and Daniel~E Ho. 2021.
\newblock When does pretraining help? assessing self-supervised learning for law and the casehold dataset of 53,000+ legal holdings.
\newblock In \emph{Proceedings of the eighteenth international conference on artificial intelligence and law}, pages 159--168.

\bibitem[{Zhong et~al.(2018)Zhong, Guo, Tu, Xiao, Liu, and Sun}]{zhong2018legal}
Haoxi Zhong, Zhipeng Guo, Cunchao Tu, Chaojun Xiao, Zhiyuan Liu, and Maosong Sun. 2018.
\newblock Legal judgment prediction via topological learning.
\newblock In \emph{Proceedings of the 2018 conference on empirical methods in natural language processing}, pages 3540--3549.

\bibitem[{Zhong et~al.(2020)Zhong, Wang, Tu, Zhang, Liu, and Sun}]{zhong2020iteratively}
Haoxi Zhong, Yuzhong Wang, Cunchao Tu, Tianyang Zhang, Zhiyuan Liu, and Maosong Sun. 2020.
\newblock Iteratively questioning and answering for interpretable legal judgment prediction.
\newblock In \emph{Proceedings of the AAAI Conference on Artificial Intelligence}, volume~34, pages 1250--1257.

\end{thebibliography}
\bibliographystyle{acl_natbib}

\appendix
\section{Implementation Details}
\label{impl}
We utilize the Adam optimizer \cite{kingma2014adam} to train our models, starting with an initial learning rate of 3e-5 for base and 3e-6 for larger models. Early stopping on validation data for up to 20 epochs is also employed. To reduce memory usage during training, we use mixed precision (fp16) and gradient accumulation. These models can handle 64 paragraphs, each with 128 tokens. The batch size is set to 8 in all base and 2 in large experiments. For MC Dropout methods, we use 10 runs. For CER, we add the regularization loss with a weight within $\{0.01, 0.05, 0.1, 0.5\}$ to choose the best one on AURCC metric. For ECE, we restrict to 10 bins. For Gamblers loss, we vary the reward within $\{1.0, 5.0, 6.5, 14.0\}$ with 4 warm-up epochs. We conduct each experiment five times with different random seeds and calculate their mean values.

\section{Detailed Experimental Results}
We provide the selective prediction and classification performance metrics for Task B in Tables \ref{tab1:taskb}, \ref{tab2:taskb}, \ref{tab3:taskb}, Task A|B in Tables \ref{tab1:taskab}, \ref{tab2:taskab}, \ref{tab3:taskab}, and Task A in Tables \ref{tab1:taska}, \ref{tab2:taska}, \ref{tab3:taska}. Across all configurations, MC Dropout methods consistently outperformed SR, with BALD leading or remaining comparable in most settings. However, in specific configurations such PV on LexLM-Base with CER, PV outperformed BALD. On training loss choice, CER demonstrated improvements over task-specific loss alone, while Gambler's loss showed competitiveness in specific configurations, surpassing even CER in instances like InCaseLawBERT-base for Task A and Task B, as well as LexLM-large for Task A|B. Interestingly, ECE lagged in most cases but outperformed CER in InCaseLawBERT-base for Task B and A|B, as well as LexLM-large for Task A|B, aligning with Gambler's superior performance in those configurations. These findings suggest a nuanced interplay between these specific models and data characteristics, prompting further investigation into their interactions and a theoretical understanding of their properties. Regarding pre-trained models, LexLM-base, particularly with certain configurations like PV and CER on Task A and B, attempted to yield better results than LexLM-large or InCaseLawBERT-base, warranting further exploration into the interplay between these techniques.

\begin{table*}[]
\begin{tabular}{|c|c|cccc|cccc|}
\hline
\multirow{2}{*}{\begin{tabular}[c]{@{}c@{}}Confidence \\ Estimator\end{tabular}} & Models & \multicolumn{4}{c|}{BERT-base}                                                                 & \multicolumn{4}{c|}{LegalBERT-base}                                                            \\ \cline{2-10} 
 & Loss   & \multicolumn{1}{c|}{AURCC} & \multicolumn{1}{c|}{RPP}   & \multicolumn{1}{c|}{Rf}    & mac.-F1 & \multicolumn{1}{c|}{AURCC} & \multicolumn{1}{c|}{RPP}   & \multicolumn{1}{c|}{Rf}    & mac.-F1 \\ \hline
SR                                                                               & Task   & \multicolumn{1}{c|}{21.94} & \multicolumn{1}{c|}{1.055} & \multicolumn{1}{c|}{0.186} & 61.34   & \multicolumn{1}{c|}{28.42} & \multicolumn{1}{c|}{1.055} & \multicolumn{1}{c|}{0.173} & 62.42   \\ \hline
SMP                                                                              & Task   & \multicolumn{1}{c|}{18.82} & \multicolumn{1}{c|}{0.915} & \multicolumn{1}{c|}{0.139} & 61.34   & \multicolumn{1}{c|}{16.49} & \multicolumn{1}{c|}{1.124} & \multicolumn{1}{c|}{0.110} & 62.42   \\ \hline
PV                                                                               & Task   & \multicolumn{1}{c|}{21.48} & \multicolumn{1}{c|}{0.909} & \multicolumn{1}{c|}{0.135} & 61.34   & \multicolumn{1}{c|}{16.36} & \multicolumn{1}{c|}{0.830} & \multicolumn{1}{c|}{0.111} & 62.42   \\ \hline
BALD                                                                             & Task   & \multicolumn{1}{c|}{17.98} & \multicolumn{1}{c|}{0.905} & \multicolumn{1}{c|}{0.134} & 61.34   & \multicolumn{1}{c|}{16.22} & \multicolumn{1}{c|}{0.831} & \multicolumn{1}{c|}{0.110} & 62.42   \\ \hline
SR                                                                               & CER    & \multicolumn{1}{c|}{22.19} & \multicolumn{1}{c|}{1.099} & \multicolumn{1}{c|}{0.209} & 62.60   & \multicolumn{1}{c|}{21.11} & \multicolumn{1}{c|}{1.035} & \multicolumn{1}{c|}{0.201} & 62.31   \\ \hline
SMP                                                                              & CER    & \multicolumn{1}{c|}{18.15} & \multicolumn{1}{c|}{0.883} & \multicolumn{1}{c|}{0.135} & 62.60   & \multicolumn{1}{c|}{16.26} & \multicolumn{1}{c|}{0.784} & \multicolumn{1}{c|}{0.125} & 62.31   \\ \hline
PV                                                                               & CER    & \multicolumn{1}{c|}{18.05} & \multicolumn{1}{c|}{0.882} & \multicolumn{1}{c|}{0.132} & 62.60   & \multicolumn{1}{c|}{16.30} & \multicolumn{1}{c|}{0.803} & \multicolumn{1}{c|}{0.127} & 62.31   \\ \hline
BALD                                                                             & CER    & \multicolumn{1}{c|}{17.92} & \multicolumn{1}{c|}{0.873} & \multicolumn{1}{c|}{0.129} & 62.60   & \multicolumn{1}{c|}{16.04} & \multicolumn{1}{c|}{0.785} & \multicolumn{1}{c|}{0.126} & 62.31   \\ \hline
SR                                                                               & ECE    & \multicolumn{1}{c|}{24.19} & \multicolumn{1}{c|}{1.165} & \multicolumn{1}{c|}{0.204} & 61.75   & \multicolumn{1}{c|}{27.67} & \multicolumn{1}{c|}{1.283} & \multicolumn{1}{c|}{0.244} & 63.32   \\ \hline
SMP                                                                              & ECE    & \multicolumn{1}{c|}{19.30} & \multicolumn{1}{c|}{0.925} & \multicolumn{1}{c|}{0.150} & 61.75   & \multicolumn{1}{c|}{19.19} & \multicolumn{1}{c|}{0.919} & \multicolumn{1}{c|}{0.217} & 63.32   \\ \hline
PV                                                                               & ECE    & \multicolumn{1}{c|}{19.09} & \multicolumn{1}{c|}{0.926} & \multicolumn{1}{c|}{0.149} & 61.75   & \multicolumn{1}{c|}{19.36} & \multicolumn{1}{c|}{0.936} & \multicolumn{1}{c|}{0.141} & 63.32   \\ \hline
BALD                                                                             & ECE    & \multicolumn{1}{c|}{18.96} & \multicolumn{1}{c|}{0.926} & \multicolumn{1}{c|}{0.147} & 61.75   & \multicolumn{1}{c|}{19.26} & \multicolumn{1}{c|}{0.921} & \multicolumn{1}{c|}{0.141} & 63.32   \\ \hline
SR                                                                               & Gamb   & \multicolumn{1}{c|}{25.21} & \multicolumn{1}{c|}{1.185} & \multicolumn{1}{c|}{0.246} & 64.32   & \multicolumn{1}{c|}{26.16} & \multicolumn{1}{c|}{1.223} & \multicolumn{1}{c|}{0.226} & 62.80   \\ \hline
SMP                                                                              & Gamb   & \multicolumn{1}{c|}{19.58} & \multicolumn{1}{c|}{0.941} & \multicolumn{1}{c|}{0.152} & 64.32   & \multicolumn{1}{c|}{17.91} & \multicolumn{1}{c|}{0.867} & \multicolumn{1}{c|}{0.132} & 62.80   \\ \hline
PV                                                                               & Gamb   & \multicolumn{1}{c|}{19.75} & \multicolumn{1}{c|}{0.959} & \multicolumn{1}{c|}{0.158} & 64.32   & \multicolumn{1}{c|}{17.94} & \multicolumn{1}{c|}{0.884} & \multicolumn{1}{c|}{0.132} & 62.80   \\ \hline
BALD                                                                             & Gamb   & \multicolumn{1}{c|}{19.64} & \multicolumn{1}{c|}{0.948} & \multicolumn{1}{c|}{0.163} & 64.32   & \multicolumn{1}{c|}{17.80} & \multicolumn{1}{c|}{0.872} & \multicolumn{1}{c|}{0.131} & 62.80   \\ \hline
\end{tabular}
\caption{Task B}
\label{tab1:taskb}
\end{table*}

\begin{table*}[]
\begin{tabular}{|c|c|cccc|cccc|}
\hline
\multirow{2}{*}{\begin{tabular}[c]{@{}c@{}}Confidence \\ Estimator\end{tabular}} & Models & \multicolumn{4}{c|}{LexLM-base}                                                                & \multicolumn{4}{c|}{LexLM-large}                                                               \\ \cline{2-10} 
& Loss   & \multicolumn{1}{c|}{AURCC} & \multicolumn{1}{c|}{RPP}   & \multicolumn{1}{c|}{Rf}    & mac.-F1 & \multicolumn{1}{c|}{AURCC} & \multicolumn{1}{c|}{RPP}   & \multicolumn{1}{c|}{Rf}    & mac.-F1 \\ \hline
SR                                                                               & Task   & \multicolumn{1}{c|}{19.33} & \multicolumn{1}{c|}{0.962} & \multicolumn{1}{c|}{0.199} & 61.42   & \multicolumn{1}{c|}{17.92} & \multicolumn{1}{c|}{0.914} & \multicolumn{1}{c|}{0.177} & 64.31   \\ \hline
SMP                                                                              & Task   & \multicolumn{1}{c|}{16.86} & \multicolumn{1}{c|}{0.835} & \multicolumn{1}{c|}{0.129} & 61.42   & \multicolumn{1}{c|}{14.06} & \multicolumn{1}{c|}{0.729} & \multicolumn{1}{c|}{0.120} & 64.31   \\ \hline
PV                                                                               & Task   & \multicolumn{1}{c|}{16.77} & \multicolumn{1}{c|}{0.840} & \multicolumn{1}{c|}{0.129} & 61.42   & \multicolumn{1}{c|}{14.04} & \multicolumn{1}{c|}{0.734} & \multicolumn{1}{c|}{0.118} & 64.31   \\ \hline
BALD                                                                             & Task   & \multicolumn{1}{c|}{16.56} & \multicolumn{1}{c|}{0.833} & \multicolumn{1}{c|}{0.129} & 61.42   & \multicolumn{1}{c|}{14.42} & \multicolumn{1}{c|}{0.745} & \multicolumn{1}{c|}{0.120} & 64.31   \\ \hline
SR                                                                               & CER    & \multicolumn{1}{c|}{20.29} & \multicolumn{1}{c|}{1.092} & \multicolumn{1}{c|}{0.198} & 65.21   & \multicolumn{1}{c|}{25.65} & \multicolumn{1}{c|}{1.178} & \multicolumn{1}{c|}{0.222} & 64.57   \\ \hline
SMP                                                                              & CER    & \multicolumn{1}{c|}{16.85} & \multicolumn{1}{c|}{0.801} & \multicolumn{1}{c|}{0.122} & 65.21   & \multicolumn{1}{c|}{14.72} & \multicolumn{1}{c|}{0.660} & \multicolumn{1}{c|}{0.112} & 64.57   \\ \hline
PV                                                                               & CER    & \multicolumn{1}{c|}{15.37} & \multicolumn{1}{c|}{0.810} & \multicolumn{1}{c|}{0.121} & 65.21   & \multicolumn{1}{c|}{14.87} & \multicolumn{1}{c|}{0.759} & \multicolumn{1}{c|}{0.112} & 64.57   \\ \hline
BALD                                                                             & CER    & \multicolumn{1}{c|}{17.06} & \multicolumn{1}{c|}{0.814} & \multicolumn{1}{c|}{0.121} & 65.21   & \multicolumn{1}{c|}{14.50} & \multicolumn{1}{c|}{0.698} & \multicolumn{1}{c|}{0.109} & 64.57   \\ \hline
SR                                                                               & ECE    & \multicolumn{1}{c|}{29.07} & \multicolumn{1}{c|}{1.336} & \multicolumn{1}{c|}{0.246} & 58.24   & \multicolumn{1}{c|}{28.78} & \multicolumn{1}{c|}{1.293} & \multicolumn{1}{c|}{0.280} & 65.82   \\ \hline
SMP                                                                              & ECE    & \multicolumn{1}{c|}{20.36} & \multicolumn{1}{c|}{0.963} & \multicolumn{1}{c|}{0.145} & 58.24   & \multicolumn{1}{c|}{16.18} & \multicolumn{1}{c|}{0.824} & \multicolumn{1}{c|}{0.131} & 65.82   \\ \hline
PV                                                                               & ECE    & \multicolumn{1}{c|}{20.29} & \multicolumn{1}{c|}{0.972} & \multicolumn{1}{c|}{0.143} & 58.24   & \multicolumn{1}{c|}{16.18} & \multicolumn{1}{c|}{0.838} & \multicolumn{1}{c|}{0.131} & 65.82   \\ \hline
BALD                                                                             & ECE    & \multicolumn{1}{c|}{19.95} & \multicolumn{1}{c|}{0.950} & \multicolumn{1}{c|}{0.141} & 58.24   & \multicolumn{1}{c|}{16.08} & \multicolumn{1}{c|}{0.836} & \multicolumn{1}{c|}{0.130} & 65.82   \\ \hline
SR                                                                               & Gamb   & \multicolumn{1}{c|}{27.06} & \multicolumn{1}{c|}{1.171} & \multicolumn{1}{c|}{0.211} & 62.70   & \multicolumn{1}{c|}{25.37} & \multicolumn{1}{c|}{1.163} & \multicolumn{1}{c|}{0.156} & 63.65   \\ \hline
SMP                                                                              & Gamb   & \multicolumn{1}{c|}{18.36} & \multicolumn{1}{c|}{0.854} & \multicolumn{1}{c|}{0.124} & 62.70   & \multicolumn{1}{c|}{15.09} & \multicolumn{1}{c|}{0.738} & \multicolumn{1}{c|}{0.124} & 63.65   \\ \hline
PV                                                                               & Gamb   & \multicolumn{1}{c|}{18.36} & \multicolumn{1}{c|}{0.856} & \multicolumn{1}{c|}{0.122} & 62.70   & \multicolumn{1}{c|}{15.26} & \multicolumn{1}{c|}{0.752} & \multicolumn{1}{c|}{0.124} & 63.65   \\ \hline
BALD                                                                             & Gamb   & \multicolumn{1}{c|}{17.87} & \multicolumn{1}{c|}{0.846} & \multicolumn{1}{c|}{0.119} & 62.70   & \multicolumn{1}{c|}{15.25} & \multicolumn{1}{c|}{0.740} & \multicolumn{1}{c|}{0.122} & 63.65   \\ \hline
\end{tabular}
\caption{Task B}
\label{tab2:taskb}
\end{table*}

\begin{table}[!ht]
\begin{tabular}{|c|c|cccc|}
\hline
\multirow{2}{*}{\begin{tabular}[c]{@{}c@{}}Confidence \\ Estimator\end{tabular}} & Models & \multicolumn{4}{c|}{InCaseLawBERT-base}                                                        \\ \cline{2-6} 
& Loss   & \multicolumn{1}{c|}{AURCC} & \multicolumn{1}{c|}{RPP}   & \multicolumn{1}{c|}{Rf}    & mac.-F1 \\ \hline
SR                                                                               & Task   & \multicolumn{1}{c|}{19.29} & \multicolumn{1}{c|}{0.961} & \multicolumn{1}{c|}{0.172} & 61.70   \\ \hline
SMP                                                                              & Task   & \multicolumn{1}{c|}{16.22} & \multicolumn{1}{c|}{0.801} & \multicolumn{1}{c|}{0.122} & 61.70   \\ \hline
PV                                                                               & Task   & \multicolumn{1}{c|}{16.31} & \multicolumn{1}{c|}{0.810} & \multicolumn{1}{c|}{0.121} & 61.70   \\ \hline
BALD                                                                             & Task   & \multicolumn{1}{c|}{16.27} & \multicolumn{1}{c|}{0.801} & \multicolumn{1}{c|}{0.119} & 61.70   \\ \hline
SR                                                                               & CER    & \multicolumn{1}{c|}{21.11} & \multicolumn{1}{c|}{1.038} & \multicolumn{1}{c|}{0.194} & 63.69   \\ \hline
SMP                                                                              & CER    & \multicolumn{1}{c|}{17.63} & \multicolumn{1}{c|}{0.678} & \multicolumn{1}{c|}{0.132} & 63.69   \\ \hline
PV                                                                               & CER    & \multicolumn{1}{c|}{17.61} & \multicolumn{1}{c|}{0.842} & \multicolumn{1}{c|}{0.130} & 63.69   \\ \hline
BALD                                                                             & CER    & \multicolumn{1}{c|}{17.55} & \multicolumn{1}{c|}{0.835} & \multicolumn{1}{c|}{0.128} & 63.69   \\ \hline
SR                                                                               & ECE    & \multicolumn{1}{c|}{23.98} & \multicolumn{1}{c|}{1.134} & \multicolumn{1}{c|}{0.208} & 58.39   \\ \hline
SMP                                                                              & ECE    & \multicolumn{1}{c|}{18.05} & \multicolumn{1}{c|}{0.872} & \multicolumn{1}{c|}{0.139} & 58.39   \\ \hline
PV                                                                               & ECE    & \multicolumn{1}{c|}{17.85} & \multicolumn{1}{c|}{0.878} & \multicolumn{1}{c|}{0.140} & 58.39   \\ \hline
BALD                                                                             & ECE    & \multicolumn{1}{c|}{17.62} & \multicolumn{1}{c|}{0.879} & \multicolumn{1}{c|}{0.141} & 58.39   \\ \hline
SR                                                                               & Gamb   & \multicolumn{1}{c|}{21.75} & \multicolumn{1}{c|}{1.059} & \multicolumn{1}{c|}{0.203} & 62.86   \\ \hline
SMP                                                                              & Gamb   & \multicolumn{1}{c|}{17.44} & \multicolumn{1}{c|}{0.834} & \multicolumn{1}{c|}{0.141} & 62.86   \\ \hline
PV                                                                               & Gamb   & \multicolumn{1}{c|}{17.61} & \multicolumn{1}{c|}{0.850} & \multicolumn{1}{c|}{0.142} & 62.86   \\ \hline
BALD                                                                             & Gamb   & \multicolumn{1}{c|}{17.52} & \multicolumn{1}{c|}{0.842} & \multicolumn{1}{c|}{0.140} & 62.86   \\ \hline
\end{tabular}
\caption{Task B}
\label{tab3:taskb}
\end{table}

\begin{table}[!ht]
\begin{tabular}{|c|c|cccc|}
\hline
\multirow{2}{*}{\begin{tabular}[c]{@{}c@{}}Confidence \\ Estimator\end{tabular}} & Model & \multicolumn{4}{c|}{InCaselawBERT-base}                                                        \\ \cline{2-6} 
& Loss  & \multicolumn{1}{c|}{AURCC} & \multicolumn{1}{c|}{RPP}   & \multicolumn{1}{c|}{Rf}    & mac.-F1 \\ \hline
SR                                                                               & Task  & \multicolumn{1}{c|}{16.37} & \multicolumn{1}{c|}{0.737} & \multicolumn{1}{c|}{0.125} & 55.99   \\ \hline
SMP                                                                              & Task  & \multicolumn{1}{c|}{9.64}  & \multicolumn{1}{c|}{0.530} & \multicolumn{1}{c|}{0.078} & 55.99   \\ \hline
PV                                                                               & Task  & \multicolumn{1}{c|}{9.79}  & \multicolumn{1}{c|}{0.543} & \multicolumn{1}{c|}{0.079} & 55.99   \\ \hline
BALD                                                                             & Task  & \multicolumn{1}{c|}{9.71}  & \multicolumn{1}{c|}{0.535} & \multicolumn{1}{c|}{0.078} & 55.99   \\ \hline
SR                                                                               & CER   & \multicolumn{1}{c|}{12.93} & \multicolumn{1}{c|}{0.665} & \multicolumn{1}{c|}{0.105} & 54.38   \\ \hline
SMP                                                                              & CER   & \multicolumn{1}{c|}{9.00}  & \multicolumn{1}{c|}{0.475} & \multicolumn{1}{c|}{0.073} & 54.38   \\ \hline
PV                                                                               & CER   & \multicolumn{1}{c|}{9.40}  & \multicolumn{1}{c|}{0.498} & \multicolumn{1}{c|}{0.075} & 54.38   \\ \hline
BALD                                                                             & CER   & \multicolumn{1}{c|}{9.81}  & \multicolumn{1}{c|}{0.524} & \multicolumn{1}{c|}{0.026} & 54.38   \\ \hline
SR                                                                               & ECE   & \multicolumn{1}{c|}{21.10} & \multicolumn{1}{c|}{0.864} & \multicolumn{1}{c|}{0.155} & 53.79   \\ \hline
SMP                                                                              & ECE   & \multicolumn{1}{c|}{10.90} & \multicolumn{1}{c|}{0.551} & \multicolumn{1}{c|}{0.092} & 53.79   \\ \hline
PV                                                                               & ECE   & \multicolumn{1}{c|}{10.97} & \multicolumn{1}{c|}{0.554} & \multicolumn{1}{c|}{0.092} & 53.79   \\ \hline
BALD                                                                             & ECE   & \multicolumn{1}{c|}{10.83} & \multicolumn{1}{c|}{0.562} & \multicolumn{1}{c|}{0.093} & 53.79   \\ \hline
SR                                                                               & Gamb  & \multicolumn{1}{c|}{24.04} & \multicolumn{1}{c|}{1.003} & \multicolumn{1}{c|}{0.190} & 56.39   \\ \hline
SMP                                                                              & Gamb  & \multicolumn{1}{c|}{13.39} & \multicolumn{1}{c|}{0.649} & \multicolumn{1}{c|}{0.103} & 56.39   \\ \hline
PV                                                                               & Gamb  & \multicolumn{1}{c|}{13.16} & \multicolumn{1}{c|}{0.642} & \multicolumn{1}{c|}{0.102} & 56.39   \\ \hline
BALD                                                                             & Gamb  & \multicolumn{1}{c|}{12.84} & \multicolumn{1}{c|}{0.645} & \multicolumn{1}{c|}{0.103} & 56.39   \\ \hline
\end{tabular}
\caption{Task A|B}
\label{tab3:taskab}
\end{table}

\begin{table*}[!]
\begin{tabular}{|c|c|cccc|cccc|}
\hline
\multirow{2}{*}{\begin{tabular}[c]{@{}c@{}}Confidence \\ Estimator\end{tabular}} & Model & \multicolumn{4}{c|}{BERT-base}                                                                 & \multicolumn{4}{c|}{LegalBERT-base}                                                            \\ \cline{2-10} 
 & Loss  & \multicolumn{1}{c|}{AURCC} & \multicolumn{1}{c|}{RPP}   & \multicolumn{1}{c|}{Rf}    & mac.-F1 & \multicolumn{1}{c|}{AURCC} & \multicolumn{1}{c|}{RPP}   & \multicolumn{1}{c|}{Rf}    & mac.-F1 \\ \hline
SR                                                                               & Task  & \multicolumn{1}{c|}{15.35} & \multicolumn{1}{c|}{0.765} & \multicolumn{1}{c|}{0.123} & 56.60   & \multicolumn{1}{c|}{18.01} & \multicolumn{1}{c|}{0.878} & \multicolumn{1}{c|}{0.126} & 57.97   \\ \hline
SMP                                                                              & Task  & \multicolumn{1}{c|}{11.39} & \multicolumn{1}{c|}{0.565} & \multicolumn{1}{c|}{0.083} & 56.60   & \multicolumn{1}{c|}{11.79} & \multicolumn{1}{c|}{0.579} & \multicolumn{1}{c|}{0.133} & 57.97   \\ \hline
PV                                                                               & Task  & \multicolumn{1}{c|}{11.33} & \multicolumn{1}{c|}{0.572} & \multicolumn{1}{c|}{0.082} & 56.60   & \multicolumn{1}{c|}{11.62} & \multicolumn{1}{c|}{0.584} & \multicolumn{1}{c|}{0.077} & 57.97   \\ \hline
BALD                                                                             & Task  & \multicolumn{1}{c|}{11.23} & \multicolumn{1}{c|}{0.570} & \multicolumn{1}{c|}{0.081} & 56.60   & \multicolumn{1}{c|}{11.45} & \multicolumn{1}{c|}{0.581} & \multicolumn{1}{c|}{0.076} & 57.97   \\ \hline
SR                                                                               & CER   & \multicolumn{1}{c|}{17.49} & \multicolumn{1}{c|}{0.817} & \multicolumn{1}{c|}{0.134} & 54.09   & \multicolumn{1}{c|}{15.01} & \multicolumn{1}{c|}{0.750} & \multicolumn{1}{c|}{0.110} & 56.36   \\ \hline
SMP                                                                              & CER   & \multicolumn{1}{c|}{11.74} & \multicolumn{1}{c|}{0.578} & \multicolumn{1}{c|}{0.079} & 54.09   & \multicolumn{1}{c|}{10.10} & \multicolumn{1}{c|}{0.512} & \multicolumn{1}{c|}{0.069} & 56.36   \\ \hline
PV                                                                               & CER   & \multicolumn{1}{c|}{11.76} & \multicolumn{1}{c|}{0.587} & \multicolumn{1}{c|}{0.078} & 54.09   & \multicolumn{1}{c|}{10.15} & \multicolumn{1}{c|}{0.525} & \multicolumn{1}{c|}{0.069} & 56.36   \\ \hline
BALD                                                                             & CER   & \multicolumn{1}{c|}{11.63} & \multicolumn{1}{c|}{0.584} & \multicolumn{1}{c|}{0.076} & 54.09   & \multicolumn{1}{c|}{10.21} & \multicolumn{1}{c|}{0.367} & \multicolumn{1}{c|}{0.069} & 56.36   \\ \hline
SR                                                                               & ECE   & \multicolumn{1}{c|}{20.27} & \multicolumn{1}{c|}{0.933} & \multicolumn{1}{c|}{0.188} & 54.76   & \multicolumn{1}{c|}{27.61} & \multicolumn{1}{c|}{1.123} & \multicolumn{1}{c|}{0.190} & 55.81   \\ \hline
SMP                                                                              & ECE   & \multicolumn{1}{c|}{12.12} & \multicolumn{1}{c|}{0.608} & \multicolumn{1}{c|}{0.112} & 54.76   & \multicolumn{1}{c|}{11.91} & \multicolumn{1}{c|}{0.614} & \multicolumn{1}{c|}{0.086} & 55.81   \\ \hline
PV                                                                               & ECE   & \multicolumn{1}{c|}{15.31} & \multicolumn{1}{c|}{0.608} & \multicolumn{1}{c|}{0.112} & 54.76   & \multicolumn{1}{c|}{12.05} & \multicolumn{1}{c|}{0.624} & \multicolumn{1}{c|}{0.086} & 55.81   \\ \hline
BALD                                                                             & ECE   & \multicolumn{1}{c|}{12.00} & \multicolumn{1}{c|}{0.599} & \multicolumn{1}{c|}{0.110} & 54.76   & \multicolumn{1}{c|}{12.10} & \multicolumn{1}{c|}{0.622} & \multicolumn{1}{c|}{0.086} & 55.81   \\ \hline
SR                                                                               & Gamb  & \multicolumn{1}{c|}{26.17} & \multicolumn{1}{c|}{0.986} & \multicolumn{1}{c|}{0.171} & 55.68   & \multicolumn{1}{c|}{28.86} & \multicolumn{1}{c|}{1.090} & \multicolumn{1}{c|}{0.225} & 56.63   \\ \hline
SMP                                                                              & Gamb  & \multicolumn{1}{c|}{14.17} & \multicolumn{1}{c|}{0.631} & \multicolumn{1}{c|}{0.102} & 55.68   & \multicolumn{1}{c|}{11.52} & \multicolumn{1}{c|}{0.573} & \multicolumn{1}{c|}{0.094} & 56.63   \\ \hline
PV                                                                               & Gamb  & \multicolumn{1}{c|}{14.13} & \multicolumn{1}{c|}{0.631} & \multicolumn{1}{c|}{0.102} & 55.68   & \multicolumn{1}{c|}{11.48} & \multicolumn{1}{c|}{0.576} & \multicolumn{1}{c|}{0.092} & 56.63   \\ \hline
BALD                                                                             & Gamb  & \multicolumn{1}{c|}{12.56} & \multicolumn{1}{c|}{0.621} & \multicolumn{1}{c|}{0.101} & 55.68   & \multicolumn{1}{c|}{11.30} & \multicolumn{1}{c|}{0.579} & \multicolumn{1}{c|}{0.091} & 56.63   \\ \hline
\end{tabular}
\caption{Task A|B}
\label{tab1:taskab}
\end{table*}

\begin{table*}[]
\begin{tabular}{|c|c|cccc|cccc|}
\hline
\multirow{2}{*}{\begin{tabular}[c]{@{}c@{}}Confidence \\ Estimator\end{tabular}} & Model & \multicolumn{4}{c|}{LexLM-base}                                                                & \multicolumn{4}{c|}{LexLM-Large}                                                               \\ \cline{2-10} 
 & Loss  & \multicolumn{1}{c|}{AURCC} & \multicolumn{1}{c|}{RPP}   & \multicolumn{1}{c|}{Rf}    & mac.-F1 & \multicolumn{1}{c|}{AURCC} & \multicolumn{1}{c|}{RPP}   & \multicolumn{1}{c|}{Rf}    & mac.-F1 \\ \hline
SR                                                                               & Task  & \multicolumn{1}{c|}{16.26} & \multicolumn{1}{c|}{0.818} & \multicolumn{1}{c|}{0.134} & 56.06   & \multicolumn{1}{c|}{22.56} & \multicolumn{1}{c|}{0.987} & \multicolumn{1}{c|}{0.162} & 59.00   \\ \hline
SMP                                                                              & Task  & \multicolumn{1}{c|}{10.86} & \multicolumn{1}{c|}{0.543} & \multicolumn{1}{c|}{0.155} & 56.06   & \multicolumn{1}{c|}{10.36} & \multicolumn{1}{c|}{0.510} & \multicolumn{1}{c|}{0.069} & 59.00   \\ \hline
PV                                                                               & Task  & \multicolumn{1}{c|}{10.81} & \multicolumn{1}{c|}{0.553} & \multicolumn{1}{c|}{0.079} & 56.06   & \multicolumn{1}{c|}{10.47} & \multicolumn{1}{c|}{0.522} & \multicolumn{1}{c|}{0.072} & 59.00   \\ \hline
BALD                                                                             & Task  & \multicolumn{1}{c|}{10.56} & \multicolumn{1}{c|}{0.538} & \multicolumn{1}{c|}{0.078} & 56.06   & \multicolumn{1}{c|}{10.57} & \multicolumn{1}{c|}{0.532} & \multicolumn{1}{c|}{0.075} & 59.00   \\ \hline
SR                                                                               & CER   & \multicolumn{1}{c|}{17.24} & \multicolumn{1}{c|}{0.852} & \multicolumn{1}{c|}{0.153} & 59.64   & \multicolumn{1}{c|}{20.86} & \multicolumn{1}{c|}{0.910} & \multicolumn{1}{c|}{0.173} & 57.10   \\ \hline
SMP                                                                              & CER   & \multicolumn{1}{c|}{9.57}  & \multicolumn{1}{c|}{0.503} & \multicolumn{1}{c|}{0.076} & 59.64   & \multicolumn{1}{c|}{8.76}  & \multicolumn{1}{c|}{0.477} & \multicolumn{1}{c|}{0.066} & 57.10   \\ \hline
PV                                                                               & CER   & \multicolumn{1}{c|}{9.65}  & \multicolumn{1}{c|}{0.510} & \multicolumn{1}{c|}{0.077} & 59.64   & \multicolumn{1}{c|}{8.83}  & \multicolumn{1}{c|}{0.488} & \multicolumn{1}{c|}{0.067} & 57.10   \\ \hline
BALD                                                                             & CER   & \multicolumn{1}{c|}{9.76}  & \multicolumn{1}{c|}{0.517} & \multicolumn{1}{c|}{0.078} & 59.64   & \multicolumn{1}{c|}{8.75}  & \multicolumn{1}{c|}{0.394} & \multicolumn{1}{c|}{0.066} & 57.10   \\ \hline
SR                                                                               & ECE   & \multicolumn{1}{c|}{25.22} & \multicolumn{1}{c|}{1.102} & \multicolumn{1}{c|}{0.189} & 54.94   & \multicolumn{1}{c|}{31.39} & \multicolumn{1}{c|}{1.207} & \multicolumn{1}{c|}{0.237} & 55.71   \\ \hline
SMP                                                                              & ECE   & \multicolumn{1}{c|}{11.24} & \multicolumn{1}{c|}{0.563} & \multicolumn{1}{c|}{0.081} & 54.94   & \multicolumn{1}{c|}{10.39} & \multicolumn{1}{c|}{0.510} & \multicolumn{1}{c|}{0.073} & 55.71   \\ \hline
PV                                                                               & ECE   & \multicolumn{1}{c|}{11.55} & \multicolumn{1}{c|}{0.587} & \multicolumn{1}{c|}{0.082} & 54.94   & \multicolumn{1}{c|}{7.93}  & \multicolumn{1}{c|}{0.523} & \multicolumn{1}{c|}{0.075} & 55.71   \\ \hline
BALD                                                                             & ECE   & \multicolumn{1}{c|}{11.39} & \multicolumn{1}{c|}{0.570} & \multicolumn{1}{c|}{0.080} & 54.94   & \multicolumn{1}{c|}{9.97}  & \multicolumn{1}{c|}{0.518} & \multicolumn{1}{c|}{0.075} & 55.71   \\ \hline
SR                                                                               & Gamb  & \multicolumn{1}{c|}{14.41} & \multicolumn{1}{c|}{0.717} & \multicolumn{1}{c|}{0.112} & 56.55   & \multicolumn{1}{c|}{27.29} & \multicolumn{1}{c|}{1.071} & \multicolumn{1}{c|}{0.210} & 58.60   \\ \hline
SMP                                                                              & Gamb  & \multicolumn{1}{c|}{10.02} & \multicolumn{1}{c|}{0.528} & \multicolumn{1}{c|}{0.077} & 56.55   & \multicolumn{1}{c|}{8.65}  & \multicolumn{1}{c|}{0.483} & \multicolumn{1}{c|}{0.075} & 58.60   \\ \hline
PV                                                                               & Gamb  & \multicolumn{1}{c|}{9.81}  & \multicolumn{1}{c|}{0.525} & \multicolumn{1}{c|}{0.076} & 56.55   & \multicolumn{1}{c|}{8.59}  & \multicolumn{1}{c|}{0.488} & \multicolumn{1}{c|}{0.075} & 58.60   \\ \hline
BALD                                                                             & Gamb  & \multicolumn{1}{c|}{9.61}  & \multicolumn{1}{c|}{0.518} & \multicolumn{1}{c|}{0.074} & 56.55   & \multicolumn{1}{c|}{8.53}  & \multicolumn{1}{c|}{0.477} & \multicolumn{1}{c|}{0.073} & 58.60   \\ \hline
\end{tabular}
\caption{Task A|B}
\label{tab2:taskab}
\end{table*}

\begin{table*}[!ht]
\begin{tabular}{|c|c|cccc|cccc|}
\hline
\multirow{2}{*}{\begin{tabular}[c]{@{}c@{}}Confidence \\ Estimator\end{tabular}} & Model & \multicolumn{4}{c|}{BERT-base}                                                                 & \multicolumn{4}{c|}{LegalBERT-base}                                                            \\ \cline{2-10} 
& Loss  & \multicolumn{1}{c|}{AURCC} & \multicolumn{1}{c|}{RPP}   & \multicolumn{1}{c|}{Rf}    & mac.-F1 & \multicolumn{1}{c|}{AURCC} & \multicolumn{1}{c|}{RPP}   & \multicolumn{1}{c|}{Rf}    & mac.-F1 \\ \hline
SR                                                                               & Task  & \multicolumn{1}{c|}{19.79} & \multicolumn{1}{c|}{0.929} & \multicolumn{1}{c|}{0.149} & 53.67   & \multicolumn{1}{c|}{15.43} & \multicolumn{1}{c|}{0.774} & \multicolumn{1}{c|}{0.148} & 59.49   \\ \hline
SMP                                                                              & Task  & \multicolumn{1}{c|}{14.28} & \multicolumn{1}{c|}{0.688} & \multicolumn{1}{c|}{0.102} & 53.67   & \multicolumn{1}{c|}{10.37} & \multicolumn{1}{c|}{0.529} & \multicolumn{1}{c|}{0.129} & 59.49   \\ \hline
PV                                                                               & Task  & \multicolumn{1}{c|}{14.14} & \multicolumn{1}{c|}{0.693} & \multicolumn{1}{c|}{0.103} & 53.67   & \multicolumn{1}{c|}{10.24} & \multicolumn{1}{c|}{0.530} & \multicolumn{1}{c|}{0.079} & 59.49   \\ \hline
BALD                                                                             & Task  & \multicolumn{1}{c|}{13.75} & \multicolumn{1}{c|}{0.670} & \multicolumn{1}{c|}{0.102} & 53.67   & \multicolumn{1}{c|}{10.13} & \multicolumn{1}{c|}{0.525} & \multicolumn{1}{c|}{0.077} & 59.49   \\ \hline
SR                                                                               & CER   & \multicolumn{1}{c|}{17.01} & \multicolumn{1}{c|}{0.788} & \multicolumn{1}{c|}{0.133} & 53.88   & \multicolumn{1}{c|}{15.62} & \multicolumn{1}{c|}{0.783} & \multicolumn{1}{c|}{0.130} & 57.29   \\ \hline
SMP                                                                              & CER   & \multicolumn{1}{c|}{13.23} & \multicolumn{1}{c|}{0.609} & \multicolumn{1}{c|}{0.095} & 53.88   & \multicolumn{1}{c|}{11.09} & \multicolumn{1}{c|}{0.554} & \multicolumn{1}{c|}{0.079} & 57.29   \\ \hline
PV                                                                               & CER   & \multicolumn{1}{c|}{13.37} & \multicolumn{1}{c|}{0.616} & \multicolumn{1}{c|}{0.094} & 53.88   & \multicolumn{1}{c|}{10.98} & \multicolumn{1}{c|}{0.555} & \multicolumn{1}{c|}{0.077} & 57.29   \\ \hline
BALD                                                                             & CER   & \multicolumn{1}{c|}{13.34} & \multicolumn{1}{c|}{0.614} & \multicolumn{1}{c|}{0.094} & 53.88   & \multicolumn{1}{c|}{10.95} & \multicolumn{1}{c|}{0.558} & \multicolumn{1}{c|}{0.077} & 57.29   \\ \hline
SR                                                                               & ECE   & \multicolumn{1}{c|}{21.47} & \multicolumn{1}{c|}{1.019} & \multicolumn{1}{c|}{0.269} & 56.24   & \multicolumn{1}{c|}{24.33} & \multicolumn{1}{c|}{1.112} & \multicolumn{1}{c|}{0.233} & 54.37   \\ \hline
SMP                                                                              & ECE   & \multicolumn{1}{c|}{14.12} & \multicolumn{1}{c|}{0.721} & \multicolumn{1}{c|}{0.112} & 56.24   & \multicolumn{1}{c|}{13.37} & \multicolumn{1}{c|}{0.675} & \multicolumn{1}{c|}{0.097} & 54.37   \\ \hline
PV                                                                               & ECE   & \multicolumn{1}{c|}{14.00} & \multicolumn{1}{c|}{0.713} & \multicolumn{1}{c|}{0.111} & 56.24   & \multicolumn{1}{c|}{20.87} & \multicolumn{1}{c|}{0.674} & \multicolumn{1}{c|}{0.094} & 54.37   \\ \hline
BALD                                                                             & ECE   & \multicolumn{1}{c|}{13.93} & \multicolumn{1}{c|}{0.709} & \multicolumn{1}{c|}{0.110} & 56.24   & \multicolumn{1}{c|}{13.00} & \multicolumn{1}{c|}{0.660} & \multicolumn{1}{c|}{0.092} & 54.37   \\ \hline
SR                                                                               & Gamb  & \multicolumn{1}{c|}{19.47} & \multicolumn{1}{c|}{0.937} & \multicolumn{1}{c|}{0.193} & 55.50   & \multicolumn{1}{c|}{24.41} & \multicolumn{1}{c|}{1.071} & \multicolumn{1}{c|}{0.193} & 54.01   \\ \hline
SMP                                                                              & Gamb  & \multicolumn{1}{c|}{12.61} & \multicolumn{1}{c|}{0.642} & \multicolumn{1}{c|}{0.111} & 55.50   & \multicolumn{1}{c|}{13.80} & \multicolumn{1}{c|}{0.672} & \multicolumn{1}{c|}{0.103} & 54.01   \\ \hline
PV                                                                               & Gamb  & \multicolumn{1}{c|}{12.46} & \multicolumn{1}{c|}{0.641} & \multicolumn{1}{c|}{0.110} & 55.50   & \multicolumn{1}{c|}{13.84} & \multicolumn{1}{c|}{0.685} & \multicolumn{1}{c|}{0.102} & 54.01   \\ \hline
BALD                                                                             & Gamb  & \multicolumn{1}{c|}{12.45} & \multicolumn{1}{c|}{0.638} & \multicolumn{1}{c|}{0.110} & 55.50   & \multicolumn{1}{c|}{13.56} & \multicolumn{1}{c|}{0.666} & \multicolumn{1}{c|}{0.100} & 54.01   \\ \hline
\end{tabular}
\caption{Task A}
\label{tab1:taska}
\end{table*}

\begin{table*}[!ht]
\begin{tabular}{|c|c|cccc|cccc|}
\hline
\multirow{2}{*}{\begin{tabular}[c]{@{}c@{}}Confidence \\ Estimator\end{tabular}} & Model & \multicolumn{4}{c|}{LexLM-base}                                                                & \multicolumn{4}{c|}{LexLM-large}                                                               \\ \cline{2-10} 
& Loss  & \multicolumn{1}{c|}{AURCC} & \multicolumn{1}{c|}{RPP}   & \multicolumn{1}{c|}{Rf}    & mac.-F1 & \multicolumn{1}{c|}{AURCC} & \multicolumn{1}{c|}{RPP}   & \multicolumn{1}{c|}{Rf}    & mac.-F1 \\ \hline
SR                                                                               & Task  & \multicolumn{1}{c|}{19.96} & \multicolumn{1}{c|}{0.931} & \multicolumn{1}{c|}{0.131} & 56.80   & \multicolumn{1}{c|}{22.83} & \multicolumn{1}{c|}{0.987} & \multicolumn{1}{c|}{0.186} & 56.37   \\ \hline
SMP                                                                              & Task  & \multicolumn{1}{c|}{8.75}  & \multicolumn{1}{c|}{0.682} & \multicolumn{1}{c|}{0.089} & 56.80   & \multicolumn{1}{c|}{10.42} & \multicolumn{1}{c|}{0.546} & \multicolumn{1}{c|}{0.076} & 56.37   \\ \hline
PV                                                                               & Task  & \multicolumn{1}{c|}{9.97}  & \multicolumn{1}{c|}{0.692} & \multicolumn{1}{c|}{0.089} & 56.80   & \multicolumn{1}{c|}{12.40} & \multicolumn{1}{c|}{0.563} & \multicolumn{1}{c|}{0.080} & 56.37   \\ \hline
BALD                                                                             & Task  & \multicolumn{1}{c|}{14.45} & \multicolumn{1}{c|}{0.683} & \multicolumn{1}{c|}{0.088} & 56.80   & \multicolumn{1}{c|}{10.32} & \multicolumn{1}{c|}{0.546} & \multicolumn{1}{c|}{0.078} & 56.37   \\ \hline
SR                                                                               & CER   & \multicolumn{1}{c|}{15.83} & \multicolumn{1}{c|}{0.287} & \multicolumn{1}{c|}{0.147} & 58.20   & \multicolumn{1}{c|}{25.98} & \multicolumn{1}{c|}{0.955} & \multicolumn{1}{c|}{0.208} & 57.08   \\ \hline
SMP                                                                              & CER   & \multicolumn{1}{c|}{11.44} & \multicolumn{1}{c|}{0.577} & \multicolumn{1}{c|}{0.078} & 58.20   & \multicolumn{1}{c|}{11.54} & \multicolumn{1}{c|}{0.474} & \multicolumn{1}{c|}{0.063} & 57.08   \\ \hline
PV                                                                               & CER   & \multicolumn{1}{c|}{8.21}  & \multicolumn{1}{c|}{0.580} & \multicolumn{1}{c|}{0.077} & 58.20   & \multicolumn{1}{c|}{9.27}  & \multicolumn{1}{c|}{0.489} & \multicolumn{1}{c|}{0.065} & 57.08   \\ \hline
BALD                                                                             & CER   & \multicolumn{1}{c|}{11.34} & \multicolumn{1}{c|}{0.580} & \multicolumn{1}{c|}{0.077} & 58.20   & \multicolumn{1}{c|}{9.16}  & \multicolumn{1}{c|}{0.483} & \multicolumn{1}{c|}{0.064} & 57.08   \\ \hline
SR                                                                               & ECE   & \multicolumn{1}{c|}{23.98} & \multicolumn{1}{c|}{0.640} & \multicolumn{1}{c|}{0.180} & 55.26   & \multicolumn{1}{c|}{27.33} & \multicolumn{1}{c|}{1.270} & \multicolumn{1}{c|}{0.203} & 58.24   \\ \hline
SMP                                                                              & ECE   & \multicolumn{1}{c|}{13.73} & \multicolumn{1}{c|}{0.678} & \multicolumn{1}{c|}{0.115} & 55.26   & \multicolumn{1}{c|}{12.33} & \multicolumn{1}{c|}{0.602} & \multicolumn{1}{c|}{0.097} & 58.24   \\ \hline
PV                                                                               & ECE   & \multicolumn{1}{c|}{13.82} & \multicolumn{1}{c|}{0.687} & \multicolumn{1}{c|}{0.112} & 55.26   & \multicolumn{1}{c|}{12.54} & \multicolumn{1}{c|}{0.619} & \multicolumn{1}{c|}{0.096} & 58.24   \\ \hline
BALD                                                                             & ECE   & \multicolumn{1}{c|}{13.76} & \multicolumn{1}{c|}{0.674} & \multicolumn{1}{c|}{0.109} & 55.26   & \multicolumn{1}{c|}{12.45} & \multicolumn{1}{c|}{0.610} & \multicolumn{1}{c|}{0.095} & 58.24   \\ \hline
SR                                                                               & Gamb  & \multicolumn{1}{c|}{27.86} & \multicolumn{1}{c|}{1.092} & \multicolumn{1}{c|}{0.186} & 55.73   & \multicolumn{1}{c|}{29.70} & \multicolumn{1}{c|}{1.120} & \multicolumn{1}{c|}{0.222} & 56.30   \\ \hline
SMP                                                                              & Gamb  & \multicolumn{1}{c|}{13.20} & \multicolumn{1}{c|}{0.591} & \multicolumn{1}{c|}{0.084} & 55.73   & \multicolumn{1}{c|}{10.55} & \multicolumn{1}{c|}{0.498} & \multicolumn{1}{c|}{0.069} & 56.30   \\ \hline
PV                                                                               & Gamb  & \multicolumn{1}{c|}{13.24} & \multicolumn{1}{c|}{0.601} & \multicolumn{1}{c|}{0.084} & 55.73   & \multicolumn{1}{c|}{10.55} & \multicolumn{1}{c|}{0.507} & \multicolumn{1}{c|}{0.070} & 56.30   \\ \hline
BALD                                                                             & Gamb  & \multicolumn{1}{c|}{11.98} & \multicolumn{1}{c|}{0.585} & \multicolumn{1}{c|}{0.083} & 55.73   & \multicolumn{1}{c|}{9.90}  & \multicolumn{1}{c|}{0.503} & \multicolumn{1}{c|}{0.069} & 56.30   \\ \hline
\end{tabular}
\caption{Task A}
\label{tab2:taska}
\end{table*}

\begin{table}[!ht]
\begin{tabular}{|c|c|cccc|}
\hline
\multirow{2}{*}{\begin{tabular}[c]{@{}c@{}}Confidence \\ Estimator\end{tabular}} & Model & \multicolumn{4}{c|}{InCaseLawBERT-base}                                                        \\ \cline{2-6} 
 & Loss  & \multicolumn{1}{c|}{AURCC} & \multicolumn{1}{c|}{RPP}   & \multicolumn{1}{c|}{Rf}    & mac.-F1 \\ \hline
SR                                                                               & Task  & \multicolumn{1}{c|}{15.04} & \multicolumn{1}{c|}{0.763} & \multicolumn{1}{c|}{0.125} & 52.43   \\ \hline
SMP                                                                              & Task  & \multicolumn{1}{c|}{11.60} & \multicolumn{1}{c|}{0.588} & \multicolumn{1}{c|}{0.090} & 52.43   \\ \hline
PV                                                                               & Task  & \multicolumn{1}{c|}{11.91} & \multicolumn{1}{c|}{0.606} & \multicolumn{1}{c|}{0.092} & 52.43   \\ \hline
BALD                                                                             & Task  & \multicolumn{1}{c|}{12.15} & \multicolumn{1}{c|}{0.610} & \multicolumn{1}{c|}{0.092} & 52.43   \\ \hline
SR                                                                               & CER   & \multicolumn{1}{c|}{15.31} & \multicolumn{1}{c|}{0.760} & \multicolumn{1}{c|}{0.147} & 54.82   \\ \hline
SMP                                                                              & CER   & \multicolumn{1}{c|}{12.01} & \multicolumn{1}{c|}{0.612} & \multicolumn{1}{c|}{0.112} & 54.82   \\ \hline
PV                                                                               & CER   & \multicolumn{1}{c|}{12.02} & \multicolumn{1}{c|}{0.617} & \multicolumn{1}{c|}{0.111} & 54.82   \\ \hline
BALD                                                                             & CER   & \multicolumn{1}{c|}{11.96} & \multicolumn{1}{c|}{0.610} & \multicolumn{1}{c|}{0.110} & 54.82   \\ \hline
SR                                                                               & ECE   & \multicolumn{1}{c|}{15.09} & \multicolumn{1}{c|}{0.764} & \multicolumn{1}{c|}{0.111} & 51.38   \\ \hline
SMP                                                                              & ECE   & \multicolumn{1}{c|}{11.30} & \multicolumn{1}{c|}{0.576} & \multicolumn{1}{c|}{0.100} & 51.38   \\ \hline
PV                                                                               & ECE   & \multicolumn{1}{c|}{11.27} & \multicolumn{1}{c|}{0.577} & \multicolumn{1}{c|}{0.100} & 51.38   \\ \hline
BALD                                                                             & ECE   & \multicolumn{1}{c|}{11.26} & \multicolumn{1}{c|}{0.582} & \multicolumn{1}{c|}{0.100} & 51.38   \\ \hline
SR                                                                               & Gamb  & \multicolumn{1}{c|}{14.90} & \multicolumn{1}{c|}{0.757} & \multicolumn{1}{c|}{0.134} & 55.24   \\ \hline
SMP                                                                              & Gamb  & \multicolumn{1}{c|}{10.43} & \multicolumn{1}{c|}{0.562} & \multicolumn{1}{c|}{0.097} & 55.24   \\ \hline
PV                                                                               & Gamb  & \multicolumn{1}{c|}{10.35} & \multicolumn{1}{c|}{0.564} & \multicolumn{1}{c|}{0.096} & 55.24   \\ \hline
BALD                                                                             & Gamb  & \multicolumn{1}{c|}{10.24} & \multicolumn{1}{c|}{0.556} & \multicolumn{1}{c|}{0.096} & 55.24   \\ \hline
\end{tabular}
\caption{Task A}
\label{tab3:taska}
\end{table}

\end{document}